\documentclass[10pt,twocolumn,letterpaper]{article}

\usepackage[pagenumbers]{iccv} % To force page numbers, e.g. for an arXiv version

\newcommand{\PAR}[1]{\vskip4pt \noindent{\bf #1~}}
\newcommand{\PARX}[1]{\vskip4pt \noindent{ #1~}}

\definecolor{iccvblue}{rgb}{0.21,0.49,0.74}
\usepackage[pagebackref,breaklinks,colorlinks,allcolors=iccvblue]{hyperref}
\usepackage{multirow}
\usepackage{xcolor}         % colors
\usepackage{colortbl}   
\usepackage{tabularx}
\usepackage{float}
\usepackage{tcolorbox}
\usepackage{makecell}

\newcommand{\highg}{\cellcolor{backgreen}}
\definecolor{deepgreen}{RGB}{0, 100, 0}
\definecolor{deepred}{RGB}{139, 0, 0}

\definecolor{backblue}{RGB}{210, 230, 250}
\newcommand{\high}{\cellcolor{backblue}}
\usepackage{xspace}
\usepackage[shortlabels]{enumitem}
\usepackage{cuted}

\definecolor{backblue}{RGB}{210, 230, 250}
\definecolor{backgreen}{RGB}{226, 240, 217}
\definecolor{backred}{RGB}{255, 223, 223}

\title{R1-Onevision: Advancing Generalized Multimodal Reasoning through Cross-Modal Formalization}
\author{
    Yi Yang$^{1^*}$
    \quad Xiaoxuan He$^{1,2^*}$
    \quad Hongkun Pan$^{1^*}$
    \quad Xiyan Jiang$^{1}$
    \quad Yan Deng$^{1}$ 
    \quad Xingtao Yang$^{1}$ \\
    \quad Haoyu Lu$^{3}$
    \quad Dacheng Yin$^{2}$
    \quad Fengyun Rao$^{2}$
    \quad Minfeng Zhu$^{1}$ 
    \quad Bo Zhang$^{1}$ 
    \quad Wei Chen$^{1}$  \\
    $^1$ Zhejiang University \quad
    $^2$ WeChat Vision, Tencent Inc.  \quad
    $^3$ Renmin University of China \quad
}

\begin{document}
% \maketitle

% \begin{strip}
%   \centering
%   \includegraphics[width=\textwidth]{fig/tisser5.pdf}
%   \captionsetup{hypcap=false}
%   \captionof{figure}{Comparison of baseline models and R1-Onevision. Deepseek-R1 struggles with perception errors due to GPT-4o’s incomplete image description and Qwen2.5-VL lacks the necessary reasoning ability to solve the problem. In contrast, R1-Onevision accurately interprets the image, applies structured reasoning, and arrives at the correct solution.}
%   \label{fig:teaser}
% \end{strip}

% \twocolumn[{
%   \centering
%   \includegraphics[width=\textwidth]{fig/tisser5.pdf}
%   \captionsetup{hypcap=false}
%   \captionof{figure}{Comparison of baseline models and R1-Onevision. Deepseek-R1 struggles with perception errors due to GPT-4o’s incomplete image description and Qwen2.5-VL lacks the necessary reasoning ability to solve the problem. In contrast, R1-Onevision accurately interprets the image, applies structured reasoning, and arrives at the correct solution.}
%   \label{fig:teaser}
%   \vspace{0.5em}
% }]

\twocolumn[{%
\renewcommand\twocolumn[1][]{#1}%
\maketitle
\vspace{-0.5em}
\includegraphics[width=\textwidth]{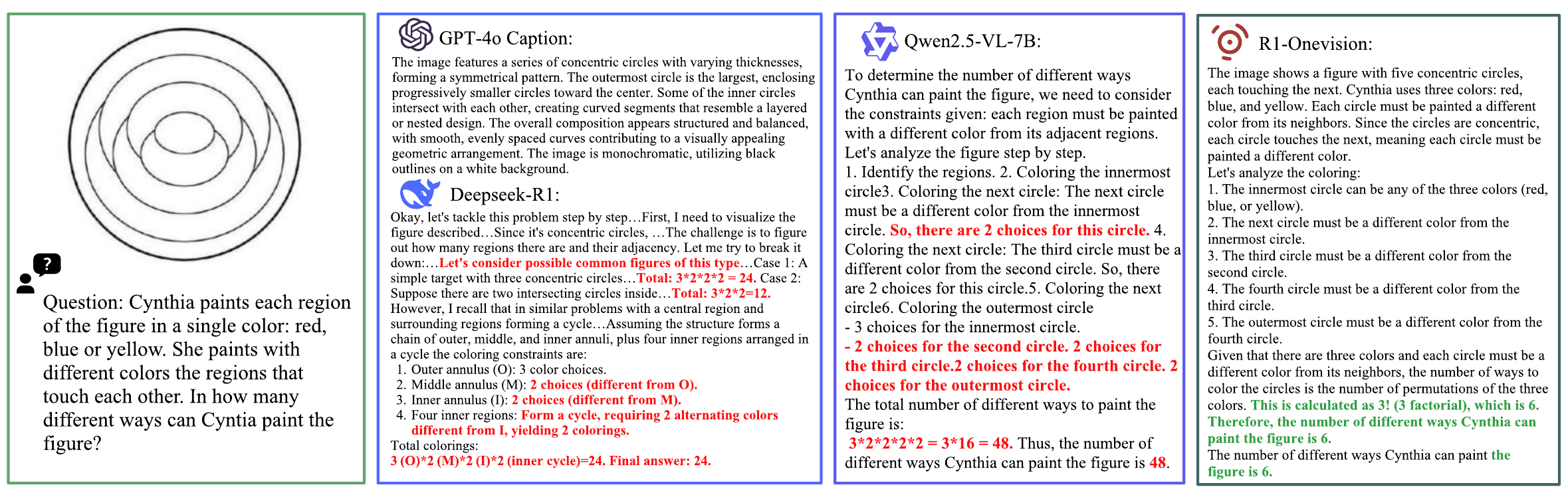}
\captionsetup{hypcap=false} 
\captionof{figure}{
Comparison of baseline models and R1-Onevision. Deepseek-R1 struggles with perception errors due to GPT-4o’s incomplete image description and Qwen2.5-VL lacks the necessary reasoning ability to solve the problem. In contrast, R1-Onevision accurately interprets the image, applies structured reasoning, and arrives at the correct solution.
}
\vspace{0.5em}
\label{fig:teaser}
}]

% \begin{figure*}[ht]
% \centering
% \includegraphics[width=\textwidth]{fig/tisser5.pdf}
% \caption{
% Comparison of baseline models and R1-Onevision. Deepseek-R1 struggles with perception errors due to GPT-4o’s incomplete image description and Qwen2.5-VL lacks the necessary reasoning ability to solve the problem. In contrast, R1-Onevision accurately interprets the image, applies structured reasoning, and arrives at the correct solution.
% }
% \label{img:teaser}
% % \vspace{-1.5em}
% \end{figure*}
% \footnote{$^*$Equal contribution}
% \begin{figure*}[!t]
%   \centering
%   \includegraphics[width=\textwidth]{fig/tisser5.pdf}
%   \captionsetup{hypcap=false}
%   \caption{Comparison of baseline models and R1-Onevision. Deepseek-R1 struggles with perception errors due to GPT-4o’s incomplete image description and Qwen2.5-VL lacks the necessary reasoning ability to solve the problem. In contrast, R1-Onevision accurately interprets the image, applies structured reasoning, and arrives at the correct solution.}
%   \label{fig:teaser}
% \end{figure*}

% \twocolumn[{%
% \renewcommand\twocolumn[1][]{#1}%
% \vspace{-0.5em}
% \includegraphics[width=\textwidth]{fig/tisser5.pdf}
% \captionsetup{hypcap=false} 
% \captionof{figure}{
% Comparison of baseline models and R1-Onevision. Deepseek-R1 struggles with perception errors due to GPT-4o’s incomplete image description and Qwen2.5-VL lacks the necessary reasoning ability to solve the problem. In contrast, R1-Onevision accurately interprets the image, applies structured reasoning, and arrives at the correct solution.
% }
% \vspace{0.5em}
% \label{fig:teaser}
% }]

\footnote{$^*$Equal contribution}

\begin{abstract}
Large Language Models have demonstrated  remarkable reasoning capability in complex textual tasks. However, multimodal reasoning, which requires integrating visual and textual information, remains a significant challenge. Existing visual-language models often struggle to effectively analyze and reason visual content, resulting in suboptimal performance on complex reasoning tasks. Moreover, the absence of comprehensive benchmarks hinders the accurate assessment of multimodal reasoning capabilities. 
In this paper, we introduce R1-Onevision, a multimodal reasoning model designed to bridge the gap between visual perception and deep reasoning. To achieve this, we propose a cross-modal reasoning pipeline that transforms images into formal textural representations, enabling precise language-based reasoning. Leveraging this pipeline, we construct the R1-Onevision dataset which provides detailed, step-by-step multimodal reasoning annotations across diverse domains. We further develop the R1-Onevision model through supervised fine-tuning and reinforcement learning to cultivate advanced reasoning and robust generalization abilities. To comprehensively evaluate multimodal reasoning performance across different grades, we introduce R1-Onevision-Bench, a benchmark aligned with human educational stages, covering exams from junior high school to university and beyond.
Experimental results show that R1-Onevision achieves state-of-the-art performance, outperforming models such as GPT-4o and Qwen2.5-VL on multiple challenging multimodal reasoning benchmarks.
 
% Large Language Models have demonstrated impressive Chain-of-Thought reasoning but enabling such complex reasoning in multimodal models remains a significant challenge. In this work, we introduce R1-Onevision, a multimodal reasoning model designed to bridge the gap between multimodal capabilities and deep reasoning abilities. To achieve this, we construct the R1-Onevision dataset by using formal language, which offers multimodal reasoning data across general, doc/chart/screen, math, and science domains. Furthermore, we conduct Supervised Fine-Tuning (SFT) on our dataset and employ Reinforcement Learning (RL) to better structure reasoning and generate reliable outputs. With its robust multimodal reasoning abilities, R1-Onevision demonstrates promising potential for applications in mathematics, science, deep image understanding, and logical reasoning. To comprehensively evaluate multimodal reasoning ability across different grades, we propose R1-Onevision-Bench, which follows human educational stages and covers tests at the junior high, high school, university, and social levels. Experimental results show that R1-Onevision achieves state-of-the-art performance, surpassing current models such as GPT-4o and Qwen2.5-VL on multiple challenging multimodal benchmarks, including R1-Onevision-Bench.    
\end{abstract}    
\section{Introduction}

% \begin{figure}
%     \centering
%     \includegraphics[width=\linewidth]{fig/tisser3.pdf}
%     \caption{}
%     \label{fig:tisser}
% \end{figure}

%第一段补充说明目前纯文本推理已经有了r1这种promising的方法。但是多模态推理目前仍然还需要探索，尝试和verify稳定有效提升推理能力的post-training方法。为了弥补这个gap，我们提出了a comprehensive post-training strategy to further boost the reasoning abilities of multimodal language models, which integrates SFT and RL. Our approach 通过teach model 如何一步步反复提取图片中信息 来完善和纠正 intermediate reasoning steps, improve logical consistency. futhermore, 通过 rewarding its formation of positive logical thinking enhance了 generalization across diverse tasks.
%GPT-o1 \cite{jaech2024openai} can accept image to reason
Recent language reasoning models like Deekseek-R1~\cite{guo2025deepseek}, chain-of-thought prompting~\cite{wei2022chain}, Cumulative Reasoning~\cite{zhang2023cumulative} have achieved remarkable progress in solving complex problems including coding~\cite{chen2021evaluating, gu2024cruxeval}, mathematics~\cite{cobbe2021training, hendrycks2024measuring}, and science~\cite{wang2024mmlu}. 
However, multimodal reasoning remains a largely underexplored challenge. Unlike textual reasoning, multimodal reasoning requires the model to iteratively extract, structure, and verify information from images.
Existing visual-language models often fail to organize available information and conduct in-depth reasoning processes, leading to failures in visual reasoning tasks. 

Current research on visual-language models has increasingly emphasized step-by-step reasoning. 
Some approaches, such as LLava-CoT~\cite{xu2025llavacotletvisionlanguage} and Llama-V-o1~\cite{thawakar2025llamav}, employ a predefined thinking structure to constrain the model’s reasoning process, limiting its robustness and creative potential. While such structured templates improve consistency, they often lead to shallow reasoning with limited comprehension.
The others, like MAmmoTH-VL~\cite{guo2024mammoth}, rely on direct imitation of curated ground-truth answers, causing models to generate responses directly without a trial-and-error manner. Consequently, these models may fail to generalize beyond their training distributions. 
An example of these limitations is shown in Fig.~\ref{fig:teaser}. Deepseek-R1 suffers from perception errors due to incomplete image descriptions from GPT-4o~\cite{openai2024gpt4ocard}, while Qwen2.5-VL~\cite{bai2025qwen2}, despite its strong multimodal capabilities, lacks deep reasoning ability and ultimately fails to solve the problem. These challenges highlight the core limitations of current multimodal reasoning approaches, which either impose rigid frameworks that restrict generalization or fail to provide human-like thinking behavior to accurately process visual information.

%without first organizing the problem and available information. 
%directly imitate the "groundtruth" answers provided in the high-quality, curated data, thus cause models often initiate responses without adequately organizing the problem and the available information. 

In addition, there is a lack of comprehensive benchmarks for evaluating multimodal reasoning capability. Existing benchmarks used by opencompass~\cite{contributors2023opencompass} mainly focus on mathematical problems, such as MathVision~\cite{wang2025measuring}, MathVista~\cite{lu2024mathvista}, and WeMath~\cite{qiao2024we}. While some more challenging datasets, like HumanEval-V~\cite{zhang2024humaneval}, Humanity’s Last Exam~\cite{phan2025humanity}, and ZeroBench~\cite{roberts2025zerobench}, aim to assess complex diagram understanding or broader knowledge proficiency. However, these benchmarks remain specialized, covering only limited aspects of reasoning.

In this work, we address these challenges by introducing a cross-modal reasoning pipeline to construct multimodal reasoning dataset, a post-training strategy to empower reasoning capabilities, and a comprehensive multimodal reasoning benchmark. 
First, we propose a \textbf{cross-modal reasoning pipeline} that transforms images into visual formal representations and allows language models to process and reason over images precisely.
We construct the \textbf{R1-Onevision dataset} to provide a detailed step-by-step multimodal reasoning process over diverse domains, including natural scenes, charts, mathematical expressions, and science.
Second, we employ a two-stage post-training strategy to train our \textbf{R1-Onevision model} with advanced reasoning abilities. The supervised fine-tuning (SFT) stage cultivates coherent thinking patterns and output structures using our R1-Onevision dataset. The reinforcement learning (RL) stage strengthen both the model’s reasoning performance and its generalization across diverse tasks. 
Third, to evaluate multimodal reasoning model, we introduce \textbf{R1-Onevision-Bench}, a comprehensive benchmark explicitly designed to evaluate ``grade-level" reasoning performance across scientific domains in the human educational system: mathematics, physics, chemistry, biology, and logical deduction.
Finally, extensive experiments on several benchmarks, including MathVision~\cite{wang2025measuring}, MathVerse~\cite{10.1007/978-3-031-73242-3_10}, MathVista~\cite{lu2024mathvista}, WeMath~\cite{qiao2024we} and our R1-Onevision-Bench, demonstrate the superior multimodal reasoning performance of our R1-Onevision model.

\PARX{Our main contributions are summarized as follows: }
% \begin{itemize} 
%     \item We present a cross-modal reasoning pipeline to construct \textbf{R1-Onevision Dataset} which encompasses a broad spectrum of images, questions and their reasoning process. 
%     \item We introduce \textbf{R1-Onevision}, a visual-language reasoning model designed for complex problem-solving.
%     \item We build \textbf{R1-Onevision-Bench} to assess multi-modal reasoning performance across different educational levels and subject areas in a course-aware manner.
%     \item Extensive experiments show the superiority of R1-Onevision compared to the baseline Qwen2.5-VL model and closed-source models like GPT-4o. 
% \end{itemize}

\begin{itemize}[label={--}]
    \item We present a cross-modal reasoning pipeline to construct \textbf{R1-Onevision Dataset} which encompasses a broad spectrum of images, questions and their reasoning process. 
    \item We introduce \textbf{R1-Onevision}, a visual-language reasoning model designed for complex problem-solving.
    \item We build \textbf{R1-Onevision-Bench} to assess multi-modal reasoning performance across different educational levels and subject areas in a course-aware manner.
    \item Extensive experiments show the superiority of R1-Onevision compared to the baseline Qwen2.5-VL model and closed-source models like GPT-4o. 
\end{itemize}

\section{Related Work}

\begin{figure*}[t]
\centering
\includegraphics[width=\textwidth]{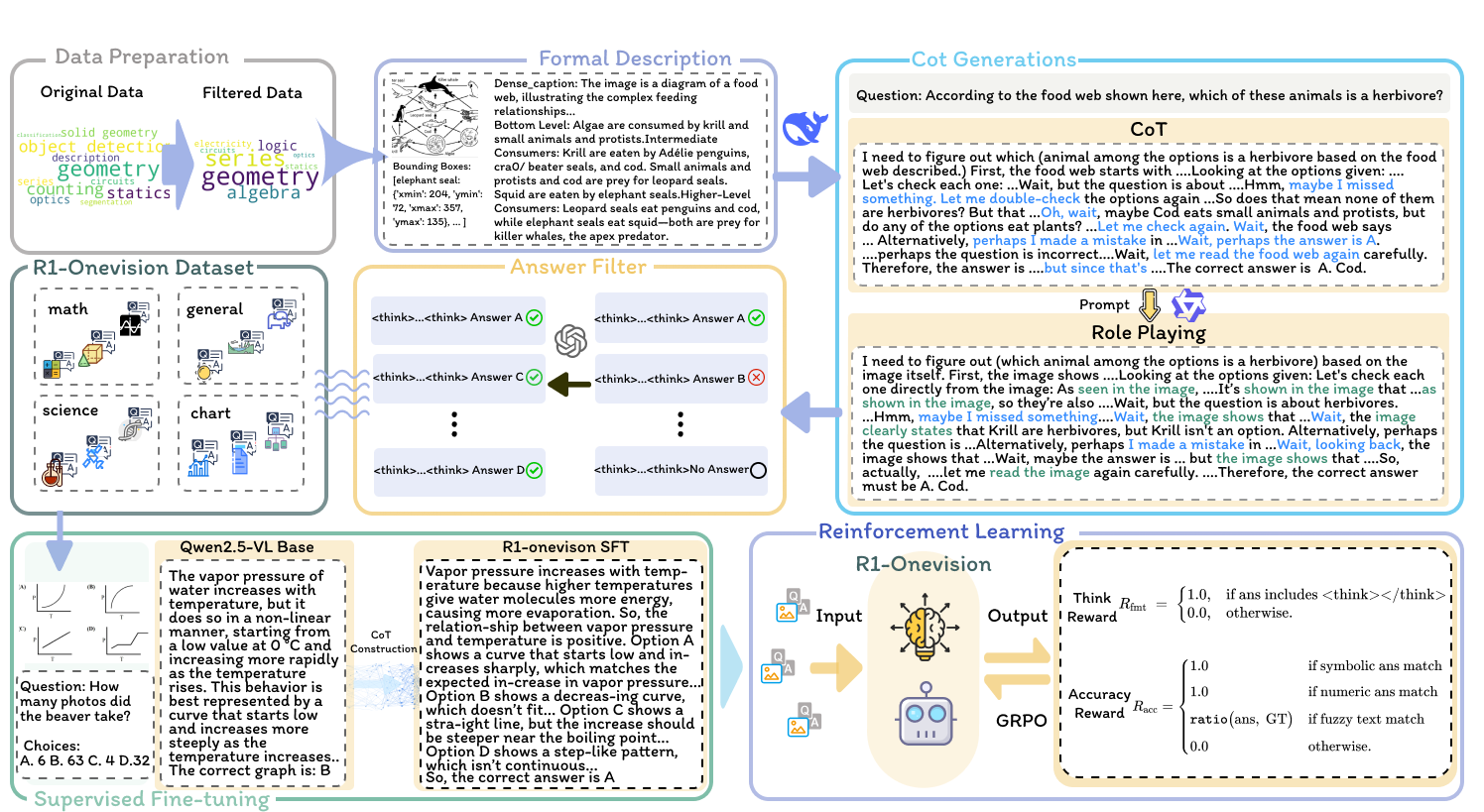}
\caption{
Overview of our R1-Onevision framework. The cross-modal reasoning pipeline begins with data preparation, integrating visual formal descriptions to enhance reasoning. We employ role-playing strategies to generate the R1-Onevision dataset with high-quality reasoning data. Post-training consists of supervised fine-Tuning to learn reasoning manner, followed by rule-based Reinforcement Learning to improve generalization across multimodal tasks.
}
\label{img:main_method}
% \vspace{-1.5em}
\end{figure*}

\PAR{Multimodal Large Language Models.} MLLMs have shown significant potential in a wide range of multimodal tasks.~\cite{meta2024llama, wang2024qwen2, bai2025qwen2, chen2024expanding} achieve excellent visual understanding ability by integrating both visual and textual data. Recently, Multimodal Large Language Models~\cite{guo2024mammoth, xu2024llava} demonstrates the advanced reasoning abilities when interpreting visual tasks. In addition, some studies~\cite{xu2024llava, thawakar2025llamav} introduce plan-based CoT prompting to guide models to generate intermediate information for predicting final answers. LLaVA-CoT~\cite{xu2024llava} introduces a novel VLM designed to conduct visual reasoning, demonstrating exceptional performance on tasks that require structured thinking and reasoning. Based on LLaVA-CoT, LlamaV-o1~\cite{thawakar2025llamav} introduces a multi-step curriculum learning approach, where tasks are progressively organized to facilitate incremental skill acquisition and problem-solving. CoMCTS~\cite{yao2024mulberry} introduces the concept of collective learning into “tree search” for effective and efficient reasoning-path searching and learning. In this paper, we propose R1-Onevision, which uses supervised finetuning and reinforcement learning during the post-training phase to generate reasoning ability to different tasks.

\PAR{Large Language Model Reasoning.} The development of robust reasoning capabilities in Large Language Models (LLMs) has been a focal point of research~\cite{huang2022towards, plaat2024reasoning, ahn2024large, kojima2022large}. Some LLMs understands and solves questions by learning to create each intermediate step of the reasoning involved till the final answer. For example, Chain-of-Thought (CoT) prompting, where a complex question is decomposed into intermediate
reasoning steps, have shown promise in guiding LLMs to structured solutions~\cite{weng2022large, yao2023tree}. Since certain visual tasks, such as solving Sudoku puzzles, urgently require multimodal large language models (MLLMs) to provide step-by-step reasoning capabilities, it is crucial to develop appropriate methods to enhance the performance of multimodal foundational models.

\PAR{Visual Reasoning Benchmarks.} With the advancement of model capabilities in visual reasoning, an increasing number of studies have proposed various benchmarks to evaluate the reasoning abilities of these models~\cite{zhang2019raven, hudson2019gqa, kahou2017figureqa}. MATH-Vision (MATH-V)~\cite{wang2025measuring} dataset collects 3,040 high-quality mathematical problems with visual contexts sourced from real math competitions. DynaMATH~\cite{zou2024dynamath} provides a dynamic visual math benchmark specifically designed to evaluate the mathematical reasoning robustness of VLMs. In this paper, we introduce a comprehensive benchmark designed to evaluate ``grade-level'' reasoning performance across scientific domains in the educational system.

\section{Method}
Multimodal reasoning requires a comprehensive understanding of both visual and textual modalities, yet existing models often fails to understand structured visual content and struggle with high-level reasoning. 
To bridge this gap, we introduce a cross-modal reasoning pipeline that transfers reasoning capabilities of language models to visual modal through visual formal representations.
Further, we employ a post-training strategy to stabilize reasoning processes and improve generalization across diverse multimodal tasks. 

\subsection{Cross-Modal Reasoning Pipeline}
Our cross-modal reasoning pipeline is designed to bridge the gap between language reasoning models and vision models by integrating visual formal representations. Formal language is a structured system with strict syntactic and semantic constraints that eliminate ambiguity, ensuring logical consistency. With formal description of visual contents, language reasoning models are able to see and reason over image elements. We use DeepSeek R1 \cite{guo2025deepseek} to generate reasoning processes on LLaVA-OneVision \cite{li2025llavaonevision} and collect these data into the R1-Onevision dataset.
% The R1-Onevision dataset is designed to advance multimodal reasoning by integrating formal language representations for images. Unlike natural language, formal language is a structured system with strict syntactic and semantic constraints that eliminate ambiguity, ensuring logical consistency. With structured annotations, the dataset precisely describes diverse domains, including natural scenes, mathematical problem-solving, and logical inference.
Figure~\ref{img:main_method} provides the process of cross-modal reasoning. Examples of the generated data are provided in the supplementary.

\PAR{Data Curation and Filtering.}
For visual reasoning, we aggregate diverse multimodal datasets covering natural images, OCR-based text extraction, charts, mathematical expressions, and scientific reasoning problems, selecting only those that support structured reasoning. The final dataset incorporates components from the LLaVA-OneVision dataset \cite{li2025llavaonevision}, augmented with domain-specific datasets tailored for complex inference tasks.
% The final dataset incorporates portions of the LLaVA-OneVision dataset \cite{}, augmented with domain-specific datasets tailored for complex multimodal inference. 

\PAR{Image Formal Description.}
A key feature of our pipeline lies in its use of formal language-based annotations. To achieve this, we leveraged a combination of GPT-4o, Grounding DINO, and EasyOCR to translate visual image content into textural formal description. We outline the annotation process below and provide details of the prompt design in the supplementary.

\begin{itemize}[label={--}]
    \item \textbf{Charts \& Diagrams}: We prompt GPT-4o to generate structured representations, such as SPICE for circuit schematics, PlantUML or Mermaid.js for flowcharts, HTML for UI layouts, CSV/JSON for tables, and Matplotlib for annotated charts. 
    \item \textbf{Natural Scenes}: We enhance images with fine-grained spatial description by leveraging Grounding DINO to extract bounding box annotations of key elements and GPT-4o to generate descriptive captions.
    \item \textbf{Text-only Images}: When dealing with images containing printed or handwritten text, we employ
    EasyOCR to extract text with positions and use GPT-4o to reconstruct the original document. 
    \item \textbf{Images with Visual and Textual Content}: We integrate GPT-4o-generated captions, Grounding DINO bounding boxes, and EasyOCR-extracted text to ensure both textual and visual elements are accurately captured.
    \item \textbf{Mathematical Images}: For images containing mathematical content, we employ GPT-4o to propose reasoning strategies to guide the inference process.
    % For images that contained mathematical content, we employed GPT-4o to summarize captions, reasoning steps, and results into dense captions for reasoning tasks.
\end{itemize}

\PAR{Reasoning Process Generation.}
Given an image, we prompt language reasoning model with its dense captions and questions to construct cross-modal Chain-of-Thought (CoT) data. While the original Chain-of-Thought (CoT) approach offered a structured reasoning path based on textual captions, it inherently lacked the essential visual component—the ability to directly ``see" and interpret the image. To address this limitation, we propose a \textbf{Role-Playing strategy} that emulates human-like visual comprehension. This method involves iteratively revisiting the image, refining its understanding, and enhancing the fidelity of the reasoning process. This process improves multimodal coherence and ensures contextually rich reasoning process.

\PAR{Quality Assurance.}
To ensure the reliability of generated reasoning process, we employ GPT-4o to remove inaccurate, irrelevant, or inconsistent CoT steps. This step guarantees a high-quality dataset for multimodal reasoning.
% Furthermore, we filtered and adjusted the data format to enhance its standardization and align it with logical reasoning.

\begin{figure}
    \centering
    \includegraphics[width=\linewidth]{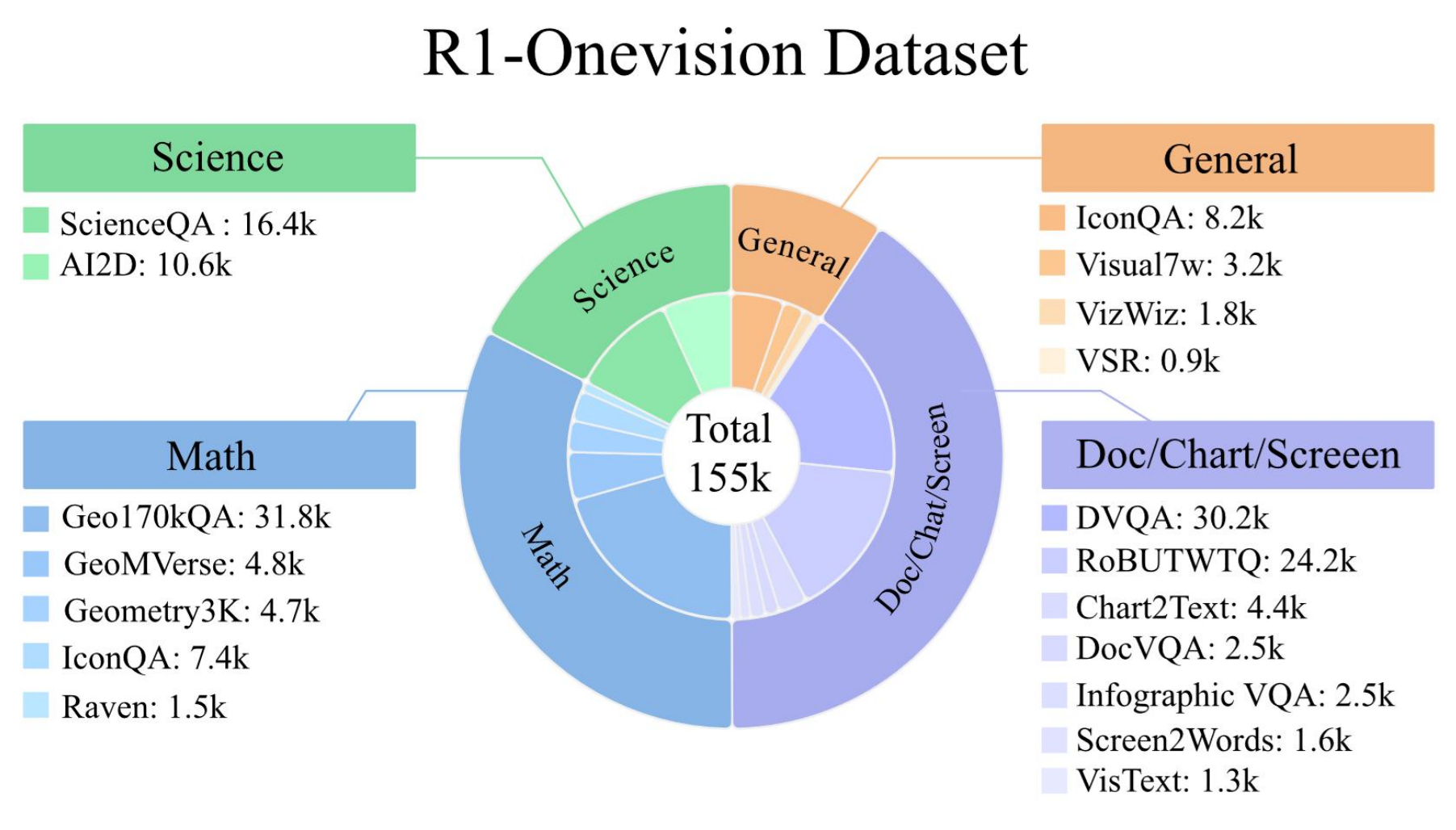}
    \caption{The data distribution of R1-Onevision dataset. R1-Onevision consists of 155k data, including Science, Math, General and Doc/Chart/Screen.}
    \label{fig:dataset}
\end{figure}

\PAR{R1-Onevision Dataset.}
Finally, as shown in the Figure~\ref{fig:dataset}, the R1-Onevision dataset is a carefully crafted tool designed to push the boundaries of multimodal reasoning. R1-Onevision encompasses a wide range of domains, including science, mathematics, chart data, and general real-world scenarios, totaling over 155k carefully curated samples. R1-Onevision serves as a rich resource for developing visual reasoning model. 

% By combining captioning techniques with formal language, reasoning methodologies, and rigorous quality control, we’ve created a dataset that not only supports reasoning tasks but also enhances the ability of models to think deeply and critically.

\subsection{Post-Training Strategy}
To enhance multimodal reasoning capabilities, we introduce a two-stage post-training strategy consisting of Supervised Fine-Tuning (SFT) and rule-based Reinforcement Learning (RL). SFT stabilizes the model’s reasoning process and standardizes its output format, while RL further improves generalization across diverse multimodal tasks.

% Our post-training strategy consists of two stages: Supervised Fine-Tuning (SFT) and rule-based Reinforcement Learning (RL). 
% In the first stage, SFT stabilizes the model's chain-of-thought outputs and cultivates a reflective reasoning process, enabling the large-scale model to handle complex problems with self-assessment. Building on the stable output format established by SFT, the subsequent RL phase further enhances the model's fundamental visual recognition capabilities, thereby improving its generalization across multi-modal tasks.

\subsubsection{Supervised Finetuning}

Leveraging the R1-Onevision dataset, we enhance the reasoning capabilities of visual language models through SFT. 
SFT stabilizes the model output format and cultivates a more sophisticated reasoning process in the large-scale model. This approach not only standardizes the output but also lays a solid foundation, enabling subsequent RL to achieve significant performance gains.

\subsubsection{Reinforcement Learning on the SFT Model}
Building upon the SFT-trained model, we employ rule-based reinforcement learning (RL) to optimize structured reasoning and ensure output validity. Specifically, we define two reward rules inspired by R1 and update the model using Group Relative Policy Optimization (GRPO). The RL stage further encourages the model to generate reliable outputs and enhances the generalizability of the model.

% Our rule-based reinforcement learning (RL) framework builds on the SFT model trained on the R1-Onevision dataset, with constraints to enforce structured reasoning and answer validity. Specifically, we develop reward rules based on R1 and update the model using Group Relative Policy Optimization (GRPO). The RL stage further refines the reasoning process to generate reliable outputs and enhances the generalizability of the model.

\PAR{Rule-Based Rewards.}
We define two reward rules that evaluate the generated answers from two perspectives:
\begin{itemize}[label={--}]
    \item Accuracy Reward: The accuracy reward rule evaluates the correctness of the final answer by extracting final answer via regular expressions and verifying them against the ground truth. For deterministic tasks such as math problems, the final answer must be provided in a specified format (e.g., within a box) to enable reliable rule-based verification. In cases like object detection, the reward is determined by the Intersection over Union (IoU) score with the ground truth.
    \item Format Reward: In order to ensure the existence of the reasoning process, the format reward rule requires that the response must follow a strict format where the model’s reasoning is enclosed between \texttt{<think>} and \texttt{</think>}. A regular expression ensures the presence and correct ordering of these reasoning markers.
\end{itemize}

% Let $r_i$ be the total reward for the i-th answer, computed by adding the correctness reward and the format reward. After collecting a batch of rewards, we often normalize them (zero mean, unit variance) to stabilize training:

% \begin{equation}
%     r_i \leftarrow \frac{r_i - \mu_r}{\sigma_r + 10^{-4}},
% \end{equation}

% where $\mu_r$ and $\sigma_r$ are the mean and standard deviation of rewards in the batch.

\PAR{Group Relative Policy Optimization.}
We employ GRPO to achieve balanced integration of consistent policy updates and robust reward signals in a controlled manner. For each token in the generated answer, GRPO first compute the log probabilities under both the new policy ($\pi(\theta)$) and a reference policy. GRPO then calculate a ratio of these probabilities and clip it to the range $[1-\epsilon, 1+\epsilon]$ to prevent overly large updates. The normalized reward (serving as the advantage) is used in a PPO-style loss:

\begin{equation}
    \mathcal{L}_{\text{clip}} = -\mathbb{E}\Bigl[\min\Bigl(\text{ratio}_t \cdot \text{Adv}_t,\ \text{clipped\_ratio}_t \cdot \text{Adv}_t\Bigr)\Bigr].  \nonumber
\end{equation}
Here, \(\text{Adv}_t\) denotes the advantage function, capturing how much better (or worse) a particular action is compared to a baseline policy value.

To further maintain closeness to the reference distribution, a KL divergence penalty (weighted by $\beta$) is added, yielding the overall loss:

\begin{equation}
\begin{split}
\mathcal{L}_{\text{GRPO}}(\theta) = -\mathbb{E}\Bigl[\, & \min\Bigl(\text{ratio}_t \cdot \text{Adv}_t,\ \text{clipped\_ratio}_t \cdot \text{Adv}_t\Bigr) \\
& - \beta \cdot \operatorname{KL}\Bigl(\pi_\theta(y\mid x),\ \pi_{\text{ref}}(y\mid x)\Bigr) \Bigr].  \nonumber
\end{split}
\end{equation}
Compared to other methods, the GRPO clipping mechanism prevents extreme policy shifts, while the KL regularization keeps the updated policy aligned with the baseline. This combination ensures that our model integrates rule-based rewards efficiently without compromising training stability.

\section{Multimodal Reasoning Benchmark}

To assess the reasoning capabilities of our R1-Onevision model, we introduce a dedicated multimodal reasoning benchmark—R1-Onevision-Bench. This benchmark is designed to comprehensively evaluate the model's performance across a wide spectrum of reasoning tasks, spanning multiple domains such as mathematics, physics, chemistry, biology, and logical deduction.

Inspired by human educational progression, R1-Onevision-Bench is structured to reflect graded levels of complexity. It includes real-world reasoning tests at the middle school, high school, and university levels, alongside a social test. This design not only mirrors the stages of cognitive development but also ensures that the evaluation covers both academic and practical reasoning skills.

\begin{table}[t]
    \centering

    % % \resizebox{1.0\columnwidth}{!}{
    \begin{tabular}{@{}l r@{}}
    %{\linewidth}{>{\raggedright\arraybackslash}X r}%{@{}l r@{}}
    \toprule
    \textbf{Statistic} & \textbf{Number} \\
    \midrule
    Total questions & 942 \\
    \quad - Multiple-choice questions & 783 (83.1\%) \\
    \quad - Free-form questions & 159 (16.9\%) \\
    \quad - \textbf{Newly collected questions} & \textbf{942 (100.0\%)} \\
    Grades/Categories/Subcategories & 4/5/38 \\
    \midrule
    Number of unique images & 938 (99.6\%) \\
    Number of unique questions & 942 (100.0\%) \\
    Number of unique answers & 152 (16.1\%) \\
    \midrule
    Maximum question length & 390 \\
    Maximum answer length & 54 \\
    Average question length & 127.2 \\
    Average answer length & 2.1 \\
    \bottomrule
    \end{tabular}
    % % }
    \caption{Key statistics of R1-Onevision-Bench.}
    \label{tab:bench_parameter}
\end{table}

\begin{figure}[ht]
    \centering
    \includegraphics[width=\linewidth]{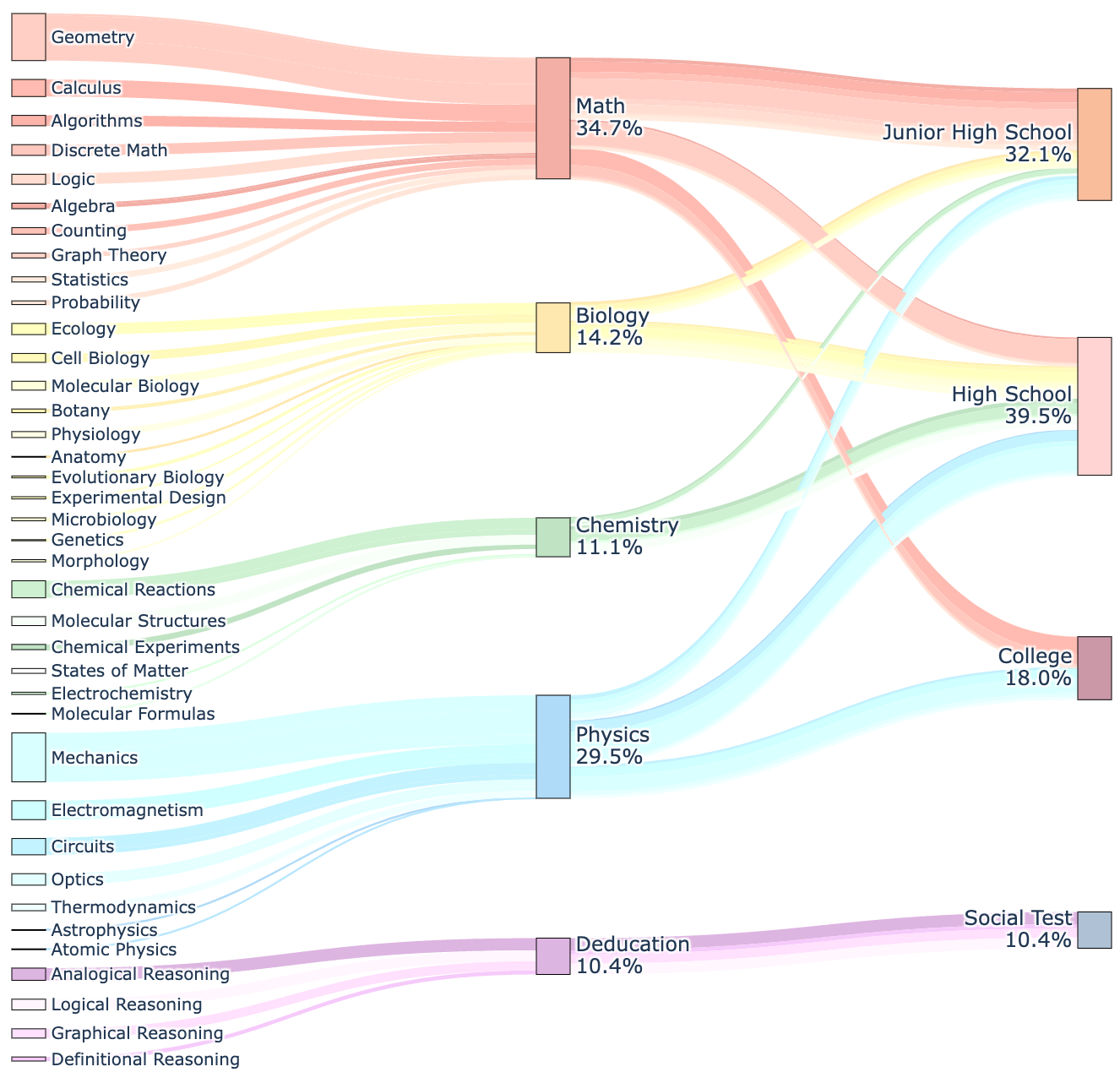}
    \caption{Overview of R1-Onevision-Bench. R1-Onevision-Bench comprises 38 subcategories organized into 5 major domains, including Math, Biology, Chemistry, Physics, Deducation. Additionally, the tasks are categorized into five levels of difficulty, ranging from `Junior High School' to `Social Test' challenges, ensuring a comprehensive evaluation of model capabilities across varying complexities.}
    \label{fig:bench_overview}
\end{figure}

\begin{figure*}[ht]
\centering
\includegraphics[width=\textwidth]{fig/bench-final.pdf}
\caption{
Samples across each category from R1-Onevision-Bench. R1-Onevision-Bench comprises multi-modal questions and answers in math, physics, chemistry, biology and deduction across multiple educational grades that require deep thinking and reasoning.
}
\label{img:bench_img}
% \vspace{-1.5em}
\end{figure*}

% \begin{figure*}[t]
% \centering
% \includegraphics[width=\textwidth]{fig/r1-bench-case.pdf}
% \caption{
% Overview of R1-Onevision-Bench. R1-Onevision-Bench comprises multi-modal questions and answers in math, physics, chemistry, biology and deduction across multiple educational grades that require deep thinking and reasoning.
% }
% \label{img:bench_img}
% % \vspace{-1.5em}
% \end{figure*}

By integrating diverse problem types and difficulty levels, R1-Onevision-Bench provides a rigorous framework for benchmarking multimodal reasoning.  This enables us to better evaluate the grade at which multimodal language models exhibit reasoning capabilities and identify which aspects of knowledge and experience require supplementation to further enhance their performance.

Moreover, our benchmark further subdivides each category into specific subcategories. Figure~\ref{img:bench_img} presents the detailed distribution of tasks across various categories, subcategories, and grade levels, along with examples of data from multiple domains.

\section{Experiment}
\begin{table*}[t]
    \footnotesize
    \centering
    \begin{tabularx}{\textwidth}{p{0.2\linewidth} *{5}{|>{\centering\arraybackslash}X}}
    %{l|c|c|c|c|c}
    \toprule[1.pt]
    \multirow{2}{*}{\textbf{Model}} & \multirow{2}{*}{\textbf{MathVision}} & \multicolumn{2}{|c|}{\textbf{MathVerse}} &  \multirow{2}{*}{\textbf{MathVista}} & \multirow{2}{*}{\textbf{WeMath}} \\ 
    \cmidrule{3-4}
    & & ALL & Vision Only & & \\
    \midrule
    \multicolumn{4}{l}{\emph{\textbf{Base Model}}} \\
    Qwen2.5-VL-7B~\cite{bai2025qwen2} & 25.4 & 43.6 & 38.2 & 63.7 & 61.0 \\
    \midrule
    \multicolumn{4}{l}{\emph{\textbf{Our Models}}} \\
    R1-Onevision-7B & 29.9 & \high{46.4} & \high{40.0} & 64.1 & 61.8 \\
    \midrule
    \multicolumn{4}{l}{\emph{\textbf{Other Models}}} \\
    GPT-4o~\cite{openai2024gpt4ocard} & \high{30.6} & 41.2 & 34.5 & 60.0 & \high{69.0}\\
    InternVL2.5-8B~\cite{chen2024expanding} & 17.1 & 35.6 & 22.8 & \high{64.5} & 53.8 \\
    Qwen2-VL-7B~\cite{wang2024qwen2} & 21.7 & 30.2 & 25.4 & 61.6 & 50.5 \\
    LLaVA-CoT-11B~\cite{xu2025llavacotletvisionlanguage} & - & - & 22.6 & 52.5 & - \\
    \bottomrule[1.pt]
    \end{tabularx}
    % }
    \caption{{Experimental results on mathematical reasoning benchmarks. Except for LLaVA-CoT-11B, all reported results are based on our reproduced experiments. The highest accuracy is marked in \colorbox{backblue}{blue}.}}
    %$^ast$ represents the reproduced result. 
    \label{tab:1}
\end{table*}

% \begin{table*}[t]
% \footnotesize
% \centering
% \setlength{\tabcolsep}{20pt}
% \begin{tabular}{l|c|c|c|c|c}
% \toprule[1.pt]
% \multirow{2}{*}{\textbf{Model}} &  \multirow{2}{*}{\textbf{MathVision}} & \multicolumn{2}{|c|}{\textbf{MathVerse}} &  \multirow{2}{*}{\textbf{MathVista}} & \multirow{2}{*}{\textbf{WeMath}} \\ 
% \cmidrule{3-4}
% & & ALL & Vision Only & & \\
% \midrule
% \multicolumn{4}{l}{\emph{\textbf{Base Model}}} \\
% Qwen2.5-VL-7B-Instruct & 25.4 & 43.58 & 38.20 & 63.7 & 60.98 \\
% \midrule
% \multicolumn{4}{l}{\emph{\textbf{Our Models}}} \\
% R1-Onevision-7B & \high{29.93} & \high{46.42} & \high{39.97} & 64.10 & \high{61.84} \\
% \midrule
% \multicolumn{4}{l}{\emph{\textbf{Other Models}}} \\

% GPT-4o & 30,59 & - & 34.5 & 60.0 & -\\
% InternVL2.5-8B & - & - & 22.8 & \high{64.5} & - \\
% Qwen2-VL-7B & - & - & 25.4 & 61.6 & - \\
% LLaVA-CoT-11B & - & - & 22.6 & 52.5 & - \\

% \bottomrule[1.pt]
% \end{tabular}
% \caption{{Experimental results on mathematical reasoning benchmarks. $^ast$ represents the reproduced result. The highest accuracy is marked in \colorbox{backblue}{blue}.}}
% \label{tab:1}
% \end{table*}

% \vspace{12em}

In this section, we first introduce the Experimental Setup of R1-Onevision in Section~\ref{sec:exp_setup}. Then, we present the main results in Section~\ref{sec:main_res}, demonstrating the effectiveness of R1-Onevision. Section~\ref{sec:bench} conduct a detailed evaluation of various models on the benchmark, including open-source and closed-source models, thoroughly assessing their performance across different grades and categories. Additionally, Section~\ref{sec:bench} provides an in-depth analysis of the areas where these models exhibit weaknesses, identifying specific challenges and limitations that hinder their effectiveness. Finally, Section~\ref{sec:analysis} conducts a systematic analysis to evaluate the importance of various components during the training process.  

\subsection{Experimental Setup}
\label{sec:exp_setup}

We evaluate R1-Onevision on several multimodal reasoning benchmarks, including MathVista~\cite{lu2024mathvista}, MathVision~\cite{wang2025measuring}, MathVerse~\cite{10.1007/978-3-031-73242-3_10} and WeMath~\cite{qiao2024we}. (1) \textbf{MathVista}: a math benchmark designed to combine challenges from diverse mathematical and visual tasks. We use the Test Mini split, around 1000 samples. (2) \textbf{MathVision}: a meticulously curated collection of 3,040 high-quality mathematical problems with visual contexts sourced from real math competitions. We use the Test Mini split, consisting of 304 samples. (3) \textbf{MathVerse}: an all-around visual math benchmark designed for an equitable and in-depth evaluation of MLLMs. We apply the full dataset and also report the `Vision-Only' result which unveils
the great challenge by rendering the whole question within the diagram. (4) \textbf{WeMath}: a benchmark designed to explore the problem-solving principles beyond the end-to-end performance. We adopt the Test Mini split of WeMath, around 1740 samples, we report the average accuracy as the main metric. The performance metrics of all baseline models are derived from VLMEvalKit’s testing results~\cite{contributors2023opencompass}.

We adopt Qwen2.5-VL series as baseline models, and conduct experiments on baselines 3B and 7B to examine the effectiveness of our method. The SFT experiments are conducted with a batch size of 128, a learning rate of 1e-5, and training over 1 epochs. Then, we perform RL on the Clevr dataset~\cite{johnson2016clevrdiagnosticdatasetcompositional}. We experiment with training subsets of 10k for a single epoch each. More details about the dataset construction, the training process and the benchmark can be found in the supplementary.

\subsection{Main Results}
\label{sec:main_res}

Table~\ref{tab:1} summarizes the quantitative results, showcasing that R1-Onevision consistently outperforms state-of-the-art (SoTA) multimodal methods across the majority of benchmarks. Notably, significant accuracy improvements are observed on MathVerse and MathVision, indicating that fine-grained visual-textual alignment and coherent chain-of-thought reasoning play a critical role in enhancing performance. These results validate the effectiveness of our proposed approach in addressing the diverse challenges of multimodal reasoning tasks. Additionally, R1-Onevision achieves 29.9\% accuracy in MathVision, which is comparable to the closed-source model GPT-4o. In MathVerse and MathVista, R1-Onevision surpasses the GPT-4o 5.2\% (MathVerse ALL), 5.5\% (MathVerse Vision Only) and 4.1\% (MathVista), further demonstrating its competitiveness in improving reasoning ability.

\begin{table*}[t]
\centering
\renewcommand{\arraystretch}{1.1} % 调整行间距

\resizebox{\textwidth}{!}{%
\begin{tabular}{l|c|cccc|ccccc}

\toprule
\multirow{2}{*}{\textbf{Model}} & \multirow{2}{*}{\textbf{Avg}} & \multicolumn{4}{|c|}{\textbf{Grade}} & \multicolumn{5}{|c}{\textbf{Category}} \\
\cmidrule{3-11}

& & Junior High School & High School & College & Social Test & Math & Physics & Chemistry & Biology & Deduction \\
\midrule
\multicolumn{11}{l}{\textbf{\textit{Closed-source}}}\\ \midrule

GPT-4o & 49.6 & 51.3 & 56.2 & 45.3 & 26.5 & 41.3 & 52.5 & 71.4 & 63.4 & 26.5 \\
Gemini-2.0-Flash & \high{59.1} & \high{56.0} & \high{65.9} & \high{61.2} & \high{39.8} & \high{52.3} & \high{64.4} & \high{74.3} & \high{67.2} & \high{39.8} \\
Claude-3.5 & 52.1 & \high{56.0} & 55.9 & 49.4 & 30.6 & 46.5 & 54.3 & 66.7 & 65.7 & 30.6 \\

\midrule
\multicolumn{11}{l}{\textbf{\textit{Open-source}}}\\ \midrule

MiniCPM-o-$2.6$~\cite{yao2024minicpm} & 30.4 & 33.4 & 31.7 & 21.2 & \highg{31.6} & 24.2 & \highg{31.7} & 30.5 & 41.8 & \highg{31.6} \\
InternVL$2.5$-8B~\cite{chen2024expanding} & 29.5 & 33.1 & 30.6 & 21.8 & 27.6 & 26.3 & 24.8 & 32.4 & 46.3 & 27.6 \\
InternVL$2.5$-8B-MPO~\cite{chen2024expanding} & 32.5 & 37.4 & 33.6 & 24.7 & 26.5 & 28.7 & 29.9 & 41.0 & 44.8 & 26.5 \\
Qwen2-VL-7B~\cite{wang2024qwen2} & 30.0 & 35.4 & 28.5 & 25.9 & 26.5 & 26.3 & 28.1 & 30.5 & 45.5 & 26.5 \\
Qwen2.5-VL-7B~\cite{bai2025qwen2} & 32.1 & 33.8 & 37.1 & 25.3 & 19.4  & 31.5 & 27.3 & 39.0 & 47.0  & 19.4 \\
DeepSeek-VL2~\cite{wu2024deepseekvl2mixtureofexpertsvisionlanguagemodels} & 29.8 & 34.4 & 30.9 & 18.8 & 30.6 & 23.5 & 28.4  & 29.5  & 47.8 & 30.6\\
R1-Onevision-7B & \highg{36.2} & \highg{40.1} & \highg{39.5} & \highg{27.6} & 26.5 & \highg{33.0} & 30.2 & \highg{49.5} & \highg{53.0} & 26.5 \\
\midrule
Qwen2.5-VL-72B~\cite{bai2025qwen2} & {52.0} & {54.3} & {56.7} & {54.1} & 23.5  & {48.9} & {55.8} & {63.8} & {63.4} & 23.5 \\
\bottomrule
\end{tabular}
}
\centering
\caption{Evaluation results of closed-source and open-source multimodal models on R1-Onevision benchmark. We divide the benchmark into four grades (`Junior Hight School', `High School', `College', `Social Test') and five categories (`Math', `Physics', `Chemistry', `Biology', `Deduction'). The highest accuracy for closed-source and open-source LMMs are marked in \colorbox{backblue}{blue} and \colorbox{backgreen}{green}, respectively.}
\label{tab:benchmark}
\end{table*}

\subsection{Benchmark Analysis}
\label{sec:bench}

We evaluate the performance of two sets of models in the R1-Onevision benchmark, which categorize into four levels of difficulty (ranging from `Junior High School' to `Social Test') and five distinct academic disciplines (Math, Physics, Chemistry, Biology, Deduction). The fist group consisted of SOTA closed-source VLMs, such as GPT-4o, Gemini-2.0-Flash and Claude-3.5. The second group comprised SOTA open-source VLMs, including MiniCPM-o-$2.6$, InternVL$2.5$-8B, InternVL$2.5$-8B-MPO, Qwen2-VL-7B, Qwen2.5-VL-7B, Qwen2.5-VL-72B, DeepSeek-VL2 and R1-Onevision. We deploy VLMEvalkit package to infer in various models. Additionally, we evaluate the impact of model parameters on benchmark performance by testing the Qwen2.5-VL
series, specifically comparing the 7B and 72B models. This analysis provides insights into how scaling model parameters influences performance across the benchmark, offering valuable guidance for future model development and optimization. As for scoring in R1-Onevision benchmark, following MathVision and MathVerse, we use GPT-4o-mini to extract the answer and score. The prompt of extract and score process is shown in the supplementary.

\noindent \textbf{Overall Results on Average Accuracy.} Table~\ref{tab:benchmark} illustrates the average performance of a variety of closed-source and open-source models. Closed-source models, such as GPT-4o, Gemini-2.0-Flash and Claude-3.5 have demonstrated excellent performance on the leaderboard, with their average scores surpassing 50\% those of other models. Notably, Gemini-2.0-Flash demonstrates robust performance across questions of varying difficulty levels, ranging from Junior High school to College-level tasks. It also excels across multiple categories, achieving scores above 50\% in math, physics, chemistry and biology. Furthermore, Gemini-2.0-Flash surpasses the closed-source model GPT-4o about 10\% average accuray, demonstrating its effectiveness in reasoning ablity. However, both closed-source and open-source models exhibit a stronger proficiency in tackling questions designed for middle and high school students, while their performance tends to decline on university-level and professional certification exams. Across various academic disciplines, all models struggle with \textbf{Deduction} questions, with none of the models exceeding 40\% accuracy. Finally, while open-source models typically underperform compared to closed-source counterparts, recent advancements in models like Qwen2.5-VL-72B has significantly narrowed the gap, with Qwen2.5-VL-72B achieves 52\% average accuracy. Qwen2.5-VL-72B is on par with close-source Claude-3.5, ranking just behind Gemini-2.0-Flash. As for the 4-10B models, notably, R1-Onevision, developed based on Qwen2.5-VL, has delivered outstanding results, ranking just behind the top closed-source models.

\subsection{Ablation Study}
\label{sec:analysis}

\subsubsection{Analysis of Training Strategy}

To evaluate the effectiveness of our training data, we compare the model's performance under two distinct training strategies: (1) applying Supervised Fine-Tuning (SFT) on our dataset, and (2) further optimizing the SFT-trained model with Reinforcement Learning (RL). As illustrated in Table~\ref{tab:training_strategies}, the application of SFT on our dataset significantly enhances the model’s performance on both the MathVision and MathVerse (Vision Only) benchmarks. Notably, the model achieves comparable results on the MathVision benchmark, demonstrating its ability to construct systematic thinking habits through our dataset. These findings underscore the value of our dataset in improving models' reasoning capabilities, particularly in tasks requiring structured and logical problem-solving. Furthermore, as shown in Table~\ref{tab:sft_rl_comparison}, the subsequent application of RL after SFT yields additional performance gains. This step pushes the model toward deeper and more deductive thinking, enabling it to tackle more complex and nuanced problems. The incremental improvements observed highlight the complementary nature of SFT and RL: while SFT establishes a robust foundation by aligning the model with high-quality reasoning patterns, RL refines and amplifies these capabilities by encouraging more advanced cognitive processes. This study demonstrates that our training data plays a pivotal role in enhancing model performance, and the combination of SFT and RL provides a powerful and effective strategy for maximizing reasoning and thinking performance. These results validate the quality and utility of our reasoning dataset.

\subsubsection{Ablation Study on Model Parameters Variants}

To demonstrate the effectiveness of our approach across models of different parameter sizes, we conducted a series of ablation experiments using Qwen2.5-VL-3B as a smaller base model. Specifically, we applied both Supervised Fine-Tuning (SFT) and Reinforcement Learning (RL) to the Qwen2.5VL-3B model and evaluated its performance. The experimental results, as summarized in Table~\ref{tab:method_comparison}, reveal significant improvements in reasoning and task performance comparing with the base Qwen2.5-VL-3B. R1-Onevision-3B achieves 23.6\% accuray in MathVision, 38.6\% in MathVerse (ALL). This performance significantly surpasses that of our base model, demonstrating a remarkable improvement in reasoning capabilities. These findings highlight the scalability and adaptability of our method, showing that it remains effective across different model sizes. 

\begin{table}[t]
% \vspace{-1.5em}
\footnotesize
\centering
% \setlength{\tabcolsep}{7.8pt}
% {
\begin{tabularx}{\linewidth}{lccc}
\toprule
\textbf{Model} &  \textbf{MathVision} & \textbf{MathVerse} & \makecell{\textbf{MathVerse} \\ \textbf{(Vision Only})} \\
%\textbf{MathVerse (Vision Only)} \\ 
\midrule
Qwen2.5-VL-7B & 25.4 & 43.6 & 38.2 \\ 
SFT & 26.3 & 43.4 & 39.7 \\ 
SFT+RL & \high{29.9} & \high{46.4} & \high{40.0} \\ 
\bottomrule
\end{tabularx}
\caption{
Ablation study on training strategies. The best results are marked in \colorbox{backblue}{blue}.
}
\label{tab:training_strategies}
% }
\end{table}

\begin{table}[t]
% \vspace{-1.5em}
\footnotesize
\centering
\setlength{\tabcolsep}{20pt}
{
\begin{tabular}{lcc}
\toprule[1.pt]
\textbf{Benchmark} & \textbf{SFT+RL} & \textbf{RL} \\
\midrule
MathVision & \high{29.9} & 28.0 \\ 
\bottomrule[1.pt]
\end{tabular}
\caption{{Comparison of MathVision performance between SFT+RL and RL. The best results are marked in \colorbox{backblue}{blue}.}}
\label{tab:sft_rl_comparison}
}
\end{table}

\begin{table}[t]
    \footnotesize
    \centering
    \begin{tabularx}{\linewidth}{lccc}
    \toprule[1.pt]
    \textbf{Model} & \textbf{MathVision} & \textbf{MathVerse} & \makecell{\textbf{MathVerse} \\ \textbf{(Vision Only})} \\ 
    \midrule
    Qwen2.5-VL-3B & 21.7 & 34.7 & 31.2 \\ 
    R1-Onevision-3B & \high{23.7} & \high{38.6} & \high{35.5} \\ 
    \bottomrule[1.pt]
    \end{tabularx}
    \caption{{Evaluation of our method using Qwen2.5-VL-3B base model on MathVision, MathVerse and MathVerse (Vision Only). The best results are marked in \colorbox{backblue}{blue}.}}
    % \caption{{Comparison of our method with different parameter values.}}
    \label{tab:method_comparison}
\end{table}
\section{Conclusion}
\label{conclusion}
\hyphenpenalty=300
\tolerance=100
In this paper, we introduce a comprehensive framework for multimodal reasoning, built upon a cross-modal formalization approach that unifies data construction, model training, and evaluation. Our framework is designed to address the inherent challenges of integrating visual and textual modalities, enabling models to reason effectively across these domains. The cross-modal reasoning pipeline at the core of this framework bridges the gap between vision and language by leveraging fine-grained alignment mechanisms and structured reasoning pathways. This framework has led to the creation of the R1-Onevision dataset, a rich resource featuring detailed step-by-step reasoning annotations designed to enhance model training and evaluation. The R1-Onevision model, trained using this framework, demonstrates strong multimodal reasoning capabilities and exhibits robust generalization across a diverse range of tasks, from visual question answering to complex problem-solving scenarios. To further support the evaluation of multimodal reasoning, we introduce R1-Onevision-Bench, a comprehensive benchmark that rigorously assesses model performance across various dimensions of reasoning. Our extensive experiments validate the effectiveness of our approach, showing significant improvements over state-of-the-art open-source models. 
%, including scientific discovery and complex decision-making tasks.
% \input{sec/7_supplement}
{
    \small
    \bibliographystyle{ieeenat_fullname}
    \bibliography{main}
}

\newpage
% \maketitle
\appendix

\section*{Roadmap of Appendix}

The structure of the appendix is delineated as follows: Descriptions of the relevant experimental details are provided in the Section~\ref{sec1}. Subsequently, Section~\ref{sec2} encompasses a presentation of supplementary visualization results.

\vspace{-0.5em}

\section{More Implementation Details}
\label{sec1}

 % 负值可以减少间距，根据需要调整数值

\subsection{Data Details}

Our cross-modal reasoning pipeline consists of: Data Curation and Filtering, Image Formal Description, Reasoning Process Generation and Quality Assurance. We adapt the following prompt in Reasoning Process Generation.

\begin{tcolorbox}[colback=white]
% \textbf{Text Summarize Prompt }: 
Answer the question and provide your reasoning process, including the following: \\
1. Simulate image reasoning: Treat the image caption as an image. Simulate reasoning by imagining you are looking at the image, and act as if you can see it. However, avoid visualization as a step in the reasoning process. \\
2. Direct visual language: Frame observations as if you are directly viewing the image (e.g., ``The image shows...").  Avoid reasoning through image caption or description. \\
3. Forbidden phrases: Avoid phrases like ``based on the caption", ``based on the description", ``visualizing the image". \\
\textbf{Question}: \textit{\{question\}} \\
\textbf{Image Content}: \textit{\{caption\}}.
\end{tcolorbox}

Then, we introduce ``role play'' to bridge the gap in in real image understanding and then filter the data. The prompts are as follows:

\begin{tcolorbox}[colback=white]
Revise the provided Chain of Thought (CoT) to follow these guidelines: \\
1. Style Shift: Convert all references to image description-based reasoning into direct image-based reasoning. For example: Replace phrases like ``based on the description'' ``based on the caption" with ``the image shows" or ``as seen in the image". \\
2. Remove image visualization step: If the CoT contains an inference step for image visualization, remove it and rewrite the CoT to reflect reasoning directly on the image itself, rather than reasoning after visualization from the image description. \\
Apply these changes rigorously to ensure that the final CoT reflects direct image interpretation, uninfluenced by description, caption, image visualization. \\
\textbf{CoT}: \textit{\{cot\}}
\end{tcolorbox}

\begin{tcolorbox}[colback=white]
Give your assistant's response. This response is the reasoning steps for the assistant to solve the problem. Please follow the following rules to evaluate whether the assistant's response is valid. \\
Rules for judging as valid: \\
1. The assistant's response has correct reasoning steps. \\
2. The assistant's response has the final reasoning answer, and the final reasoning answer is consistent with the meaning of the standard answer.\\
3. The assistant's response is based on the reasoning process of the image, not the image description or caption.\\
4. There are no steps in the assistant's response that are irrelevant to the reasoning, and each reasoning step is closely related.\\
\textbf{Standard answer}: \textit{\{gt\}} \\
\textbf{Assistant's response}: \textit{\{augmented answer\}}\\
\textbf{Output}: \\
% Judgment result: <valid or invalid> \\
% Judgment basis: <If it is judged to be valid, briefly explain why the assistant’s response follows the above four rules. If it is judged to be invalid, briefly explain which rule the assistant’s response does not follow.
\end{tcolorbox}

\subsection{Model Details}
For model training, we utilized the llama-factory and adopted a full fine-tuning startegy to optimize the model's performance. Following this, we further refined the model through Reinforcement Learning (RL). During the inference phase, the RL-tuned model was evaluated using a fixed prompt to ensure consistent and reliable results. The prompt is defined as: \textbf{``First output the thinking process in $<$think$>$ $<$/think$>$ tags and then output the final answer in $<$answer$>$ $<$/answer$>$ tags.''}.

\subsection{Evaluation Details}
To evaluate the reasoning capabilities of the models, we adopted a unified evaluation framework for both public benchmarks and our benchmark R1-Onevision bench. Following~\cite{lu2024mathvista, wang2025measuring}, we utilized GPT-4o-mini~\cite{achiam2023gpt} to assess model performance. For extracting answers from Chain-of-Thought (CoT) responses, we employed the prompt \textbf{``Below is a thought process and an answer that includes the final choices. Please extract only the final choices (A, B, C, D, etc.) from the text and do not return any other words. If the final choice is not explicitly stated in the text, output NONE. No reasoning is required; just extract the answer.''} for multiple choice questions and \textbf{``The following is a thought process and a free-form answer. Please extract the numerical value or text of the final answer, excluding any explanation. If the final answer cannot be extracted from the given text, output NONE. No reasoning is required; just extract the answer.''} for free form questions. Then we score the extracted answer with the groundtruth using \textbf{``Compare `final answer' with `groundtruth'. If final answer matches `groundtruth', output YES; otherwise, output NO. Do not return any extra words. For numerical answers with units, if `final answer' contains the unit but its numeric value matches `groundtruth', consider it a match.''} 

\vspace{-1em}

\section{Visualization}
\label{sec2}

\subsection{Image Formal Description}

In this section, we demonstrate some examples of image formal description.

\begin{figure}[H]
    \centering
    \includegraphics[width=0.9\linewidth]{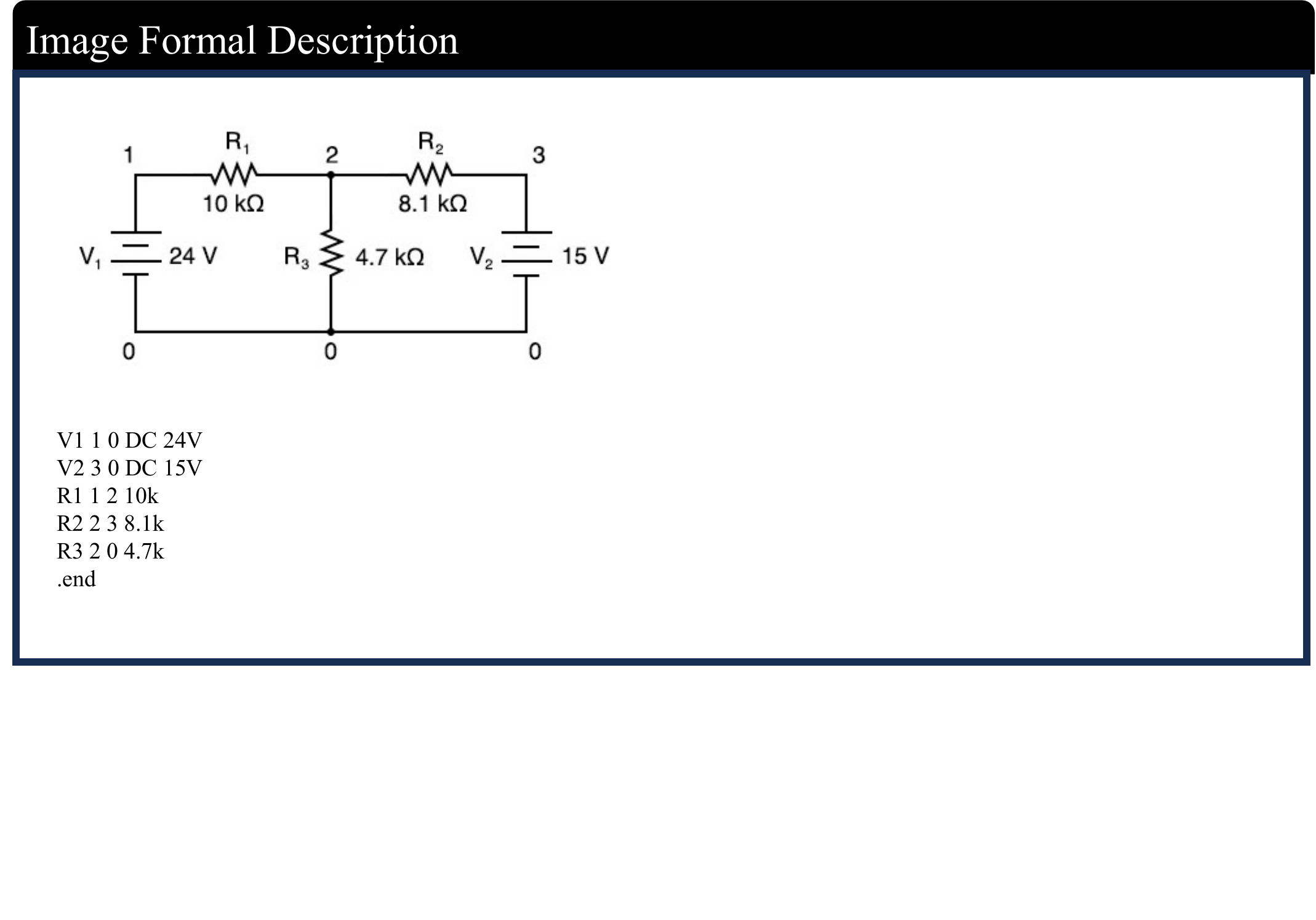}
    \caption{Formal Description of Circuts.}
    \label{fig:app1}
\end{figure}

\begin{figure}[H]
    \centering
    \includegraphics[width=0.9\linewidth]{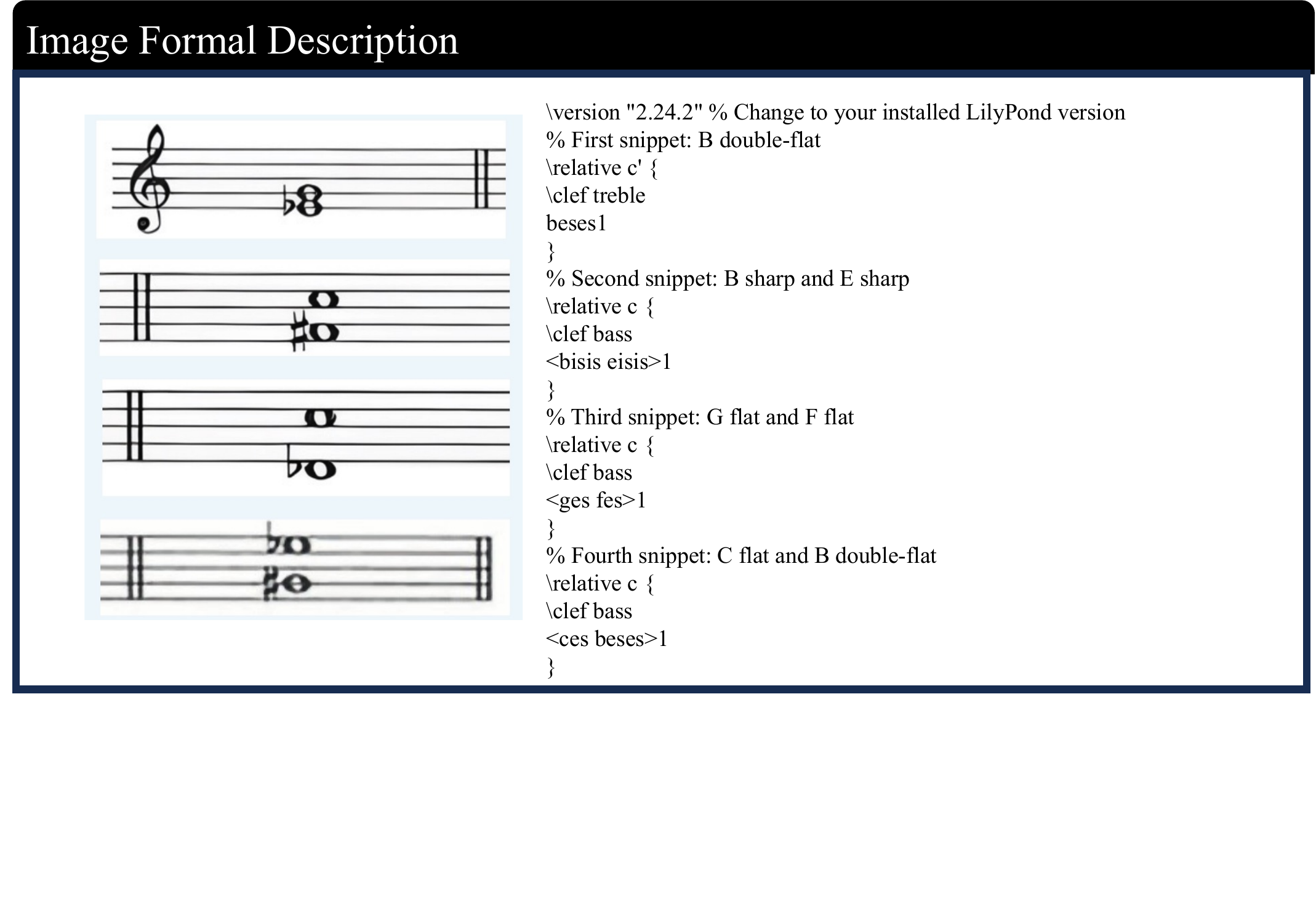}
    \caption{Formal Description of Music Sheet.}
    \label{fig:app2}
\end{figure}

\begin{figure}[H]
    \centering
    \includegraphics[width=0.9\linewidth]{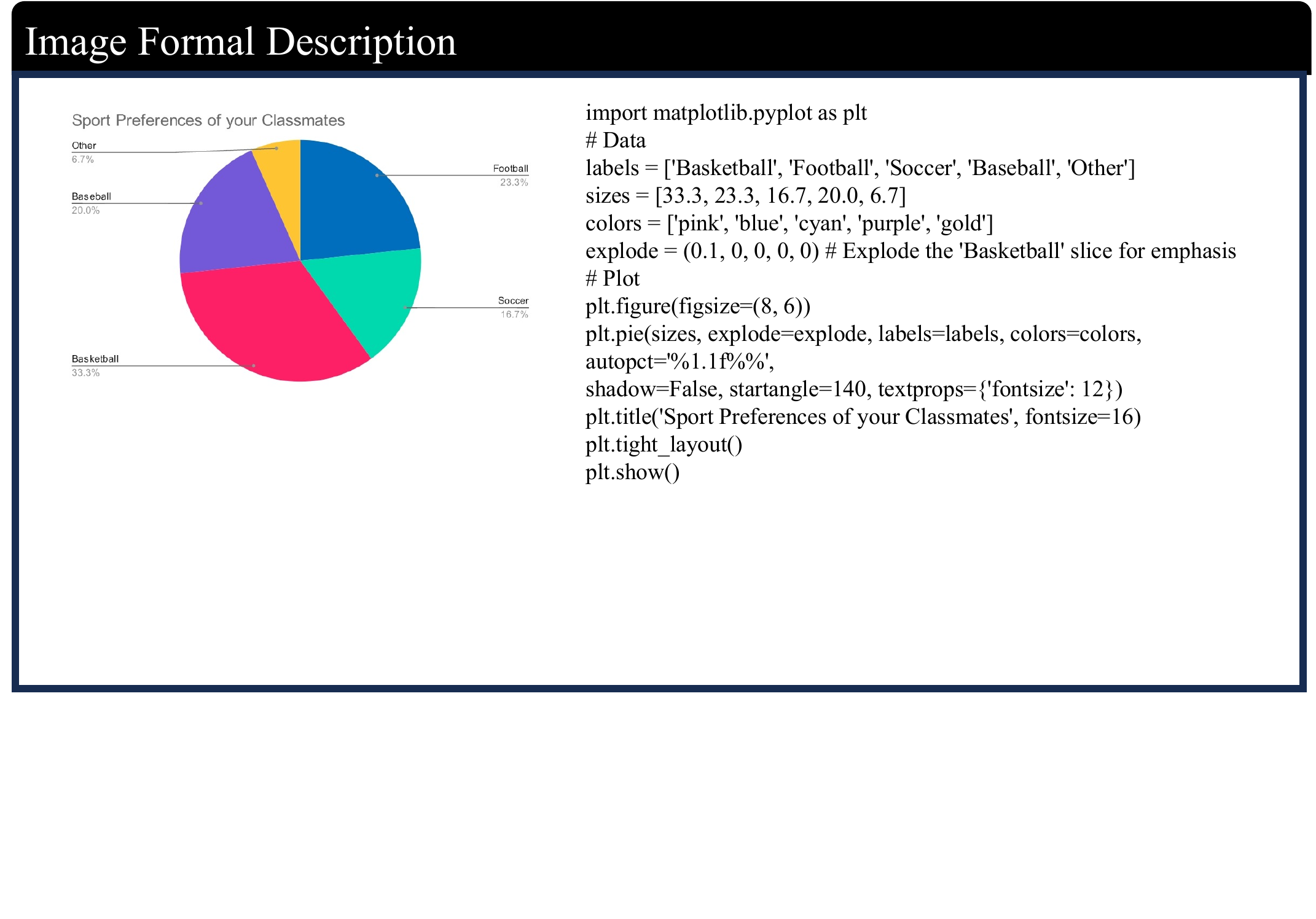}
    \caption{Formal Description of Table.}
    \label{fig:app3}
\end{figure}

\begin{figure}[H]
    \centering
    \includegraphics[width=0.9\linewidth]{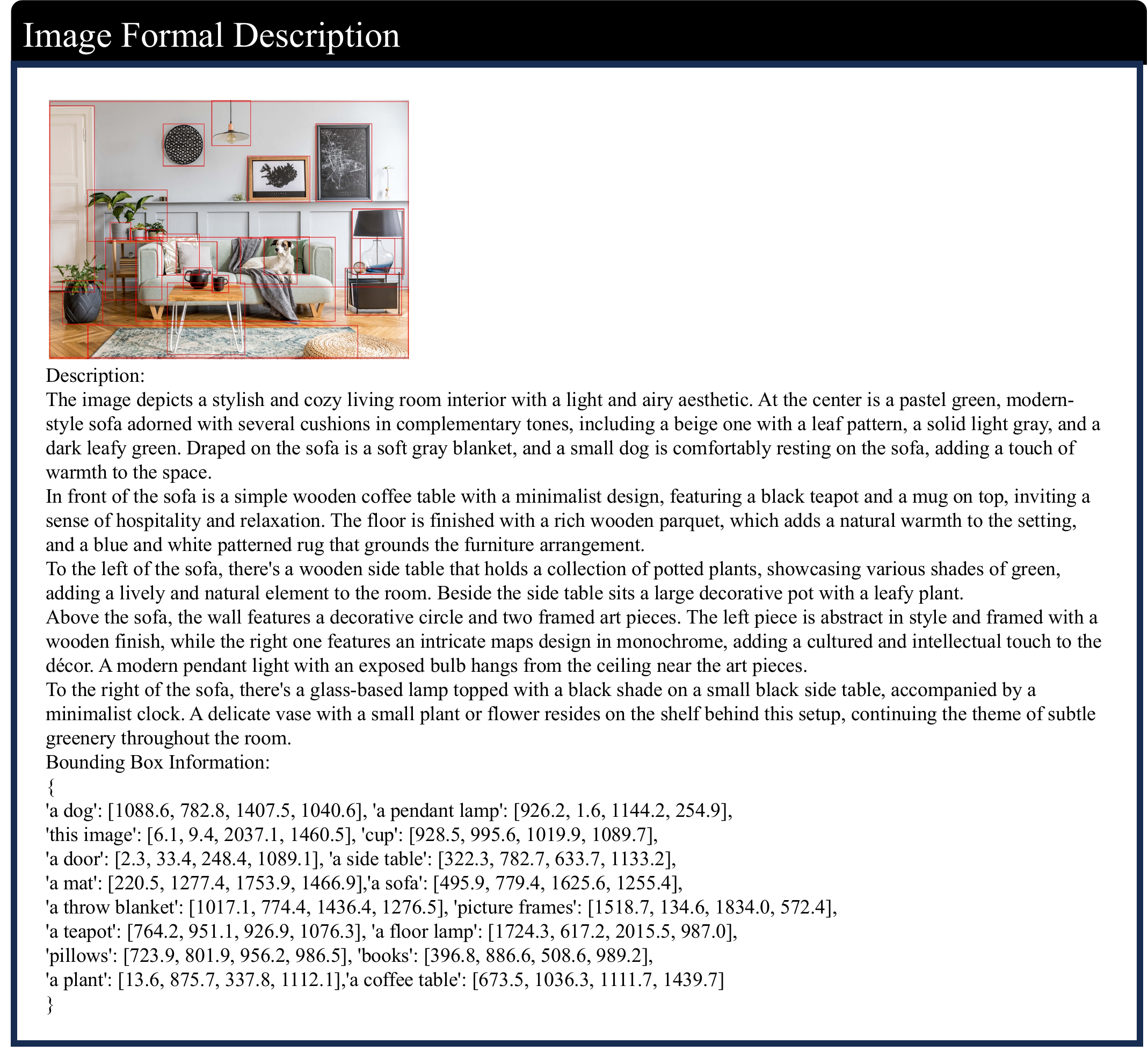}
    %[width=0.9\linewidth]{fig/4.pdf}
    \caption{Formal Description of Natural Scene.}
    \label{fig:app4}
\end{figure}

\begin{figure}[H]
    \centering
    \includegraphics[width=0.9\linewidth]{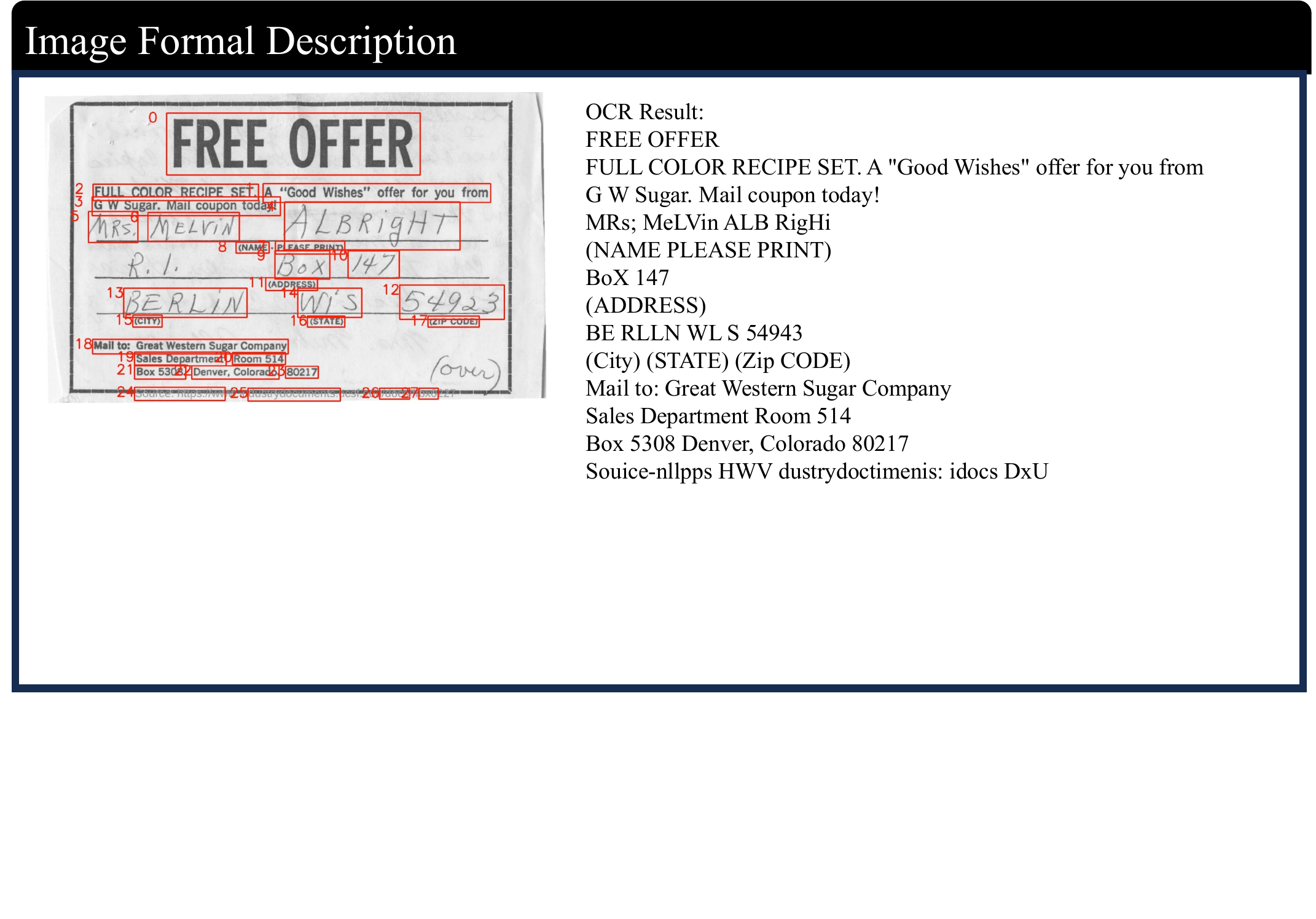}
    \caption{Formal Description of print, handwrite.}
    \label{fig:app5}
\end{figure}

\begin{figure}[H]
    \centering
    \includegraphics[width=0.9\linewidth]{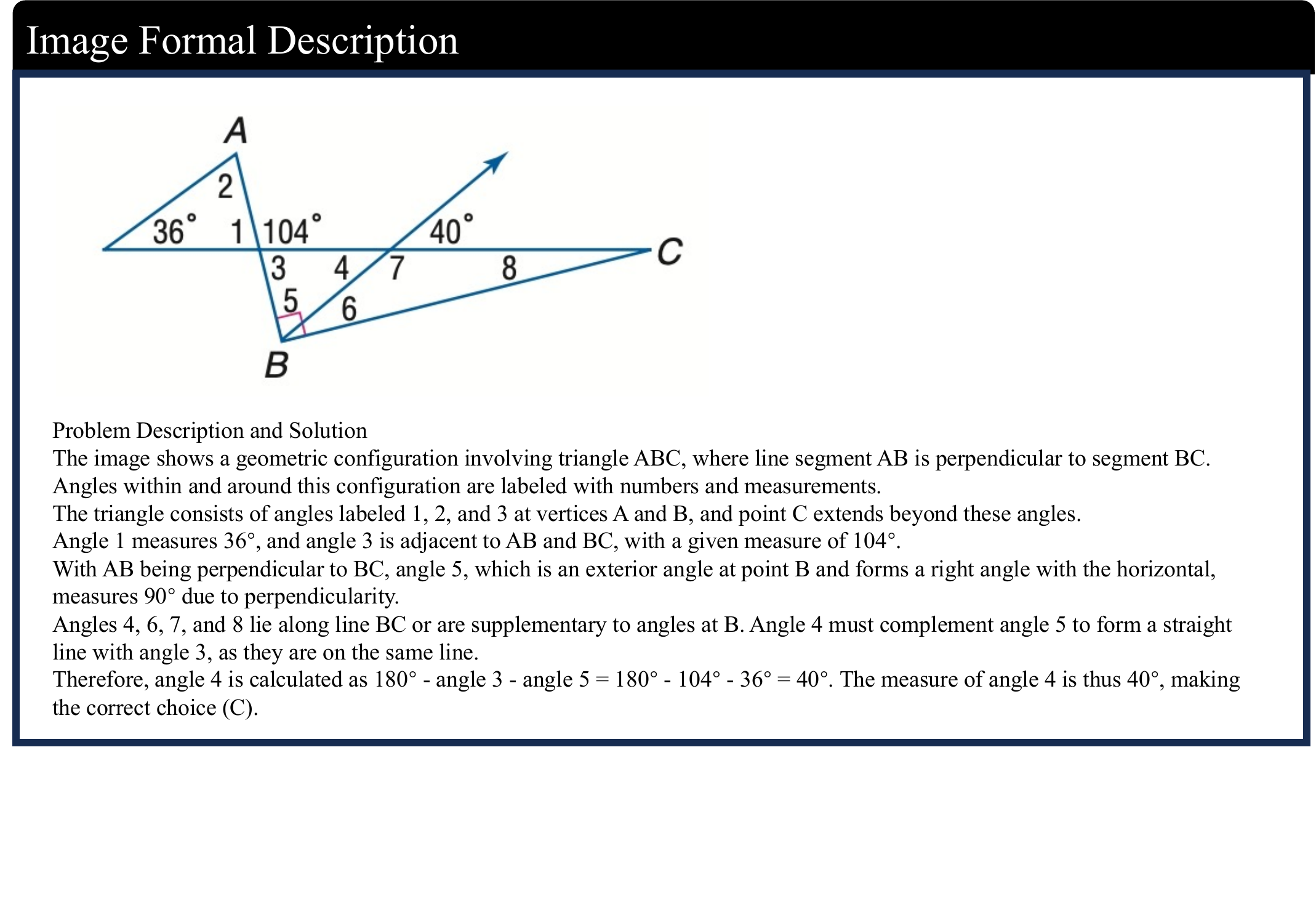}
    \caption{Formal Description of Math.}
    \label{fig:app6}
\end{figure}

\begin{figure}[H]
    \centering
    \includegraphics[width=0.9\linewidth]{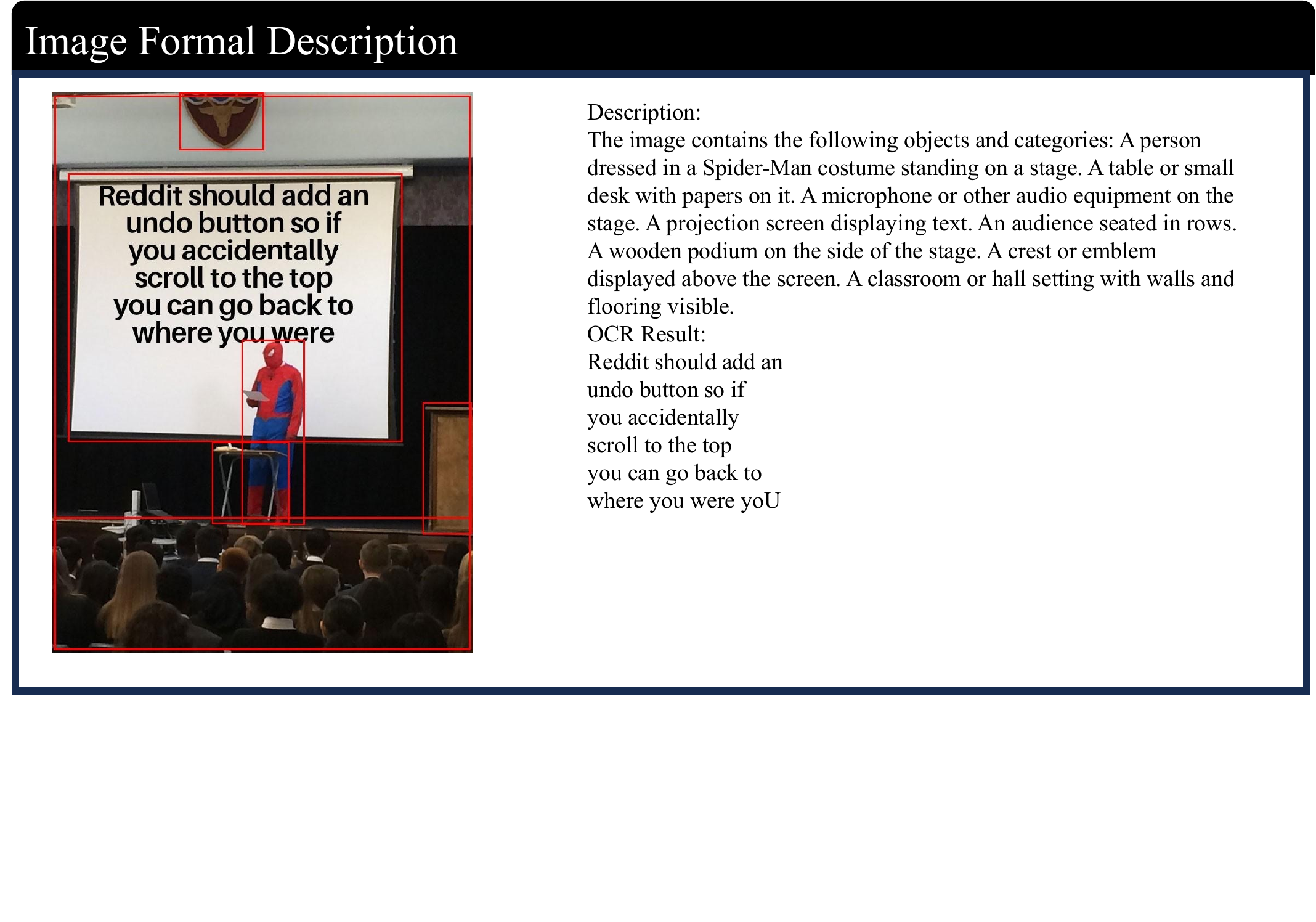}
    \caption{Formal Description of MEME.}
    \label{fig:app7}
\end{figure}

\vspace{-0.75em}

\subsection{Model Reasoning Capabilities and Qualitative Examples}

In this section, we present some practical examples of our R1-Onevision.

\begin{figure}[H]
    \centering
    \includegraphics[width=0.9\linewidth]{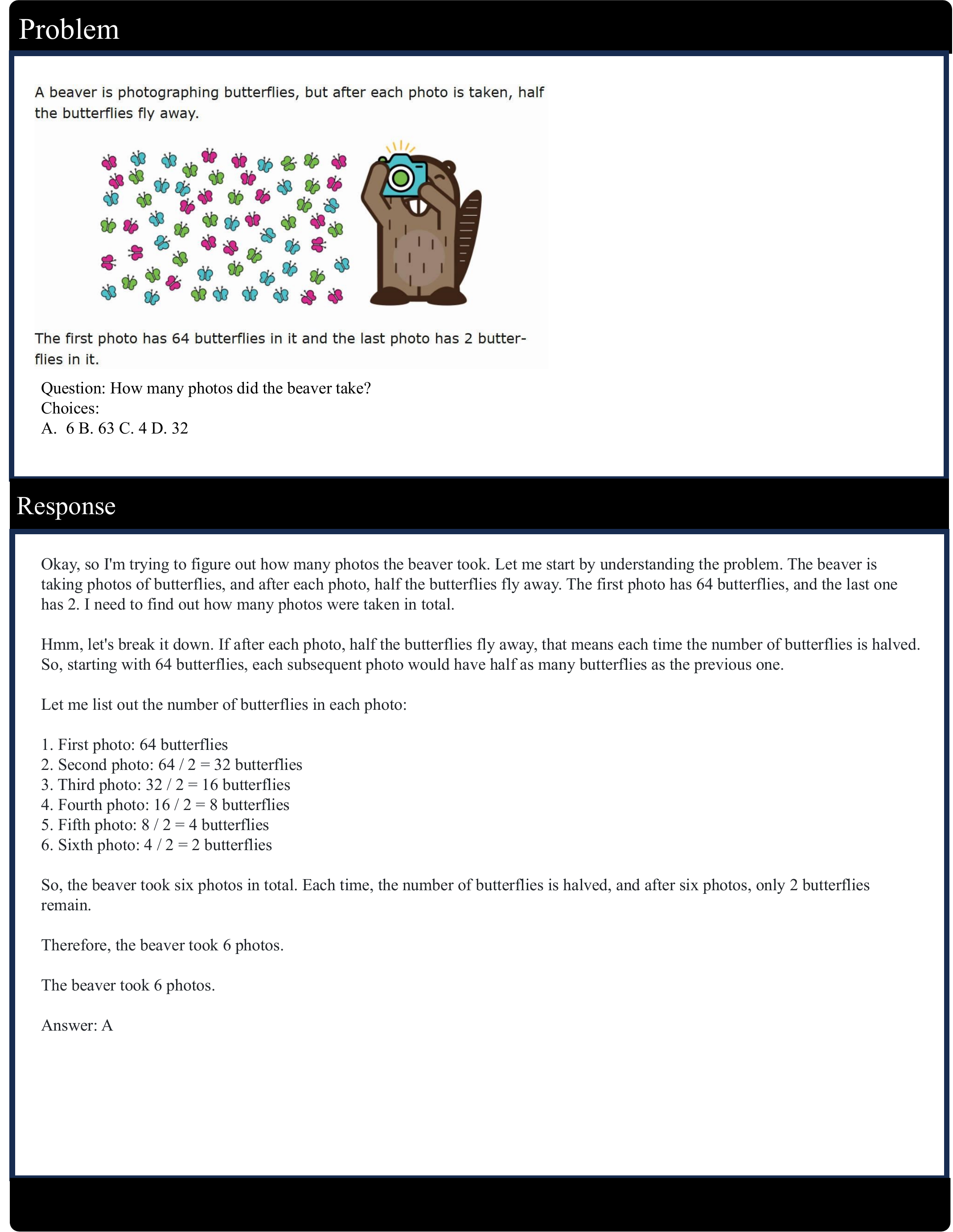}
    \caption{The model displays an adeptness in processing math.}
    \label{fig:appendix1}
\end{figure}

\begin{figure}[H]
    \centering
    \includegraphics[width=0.9\linewidth]{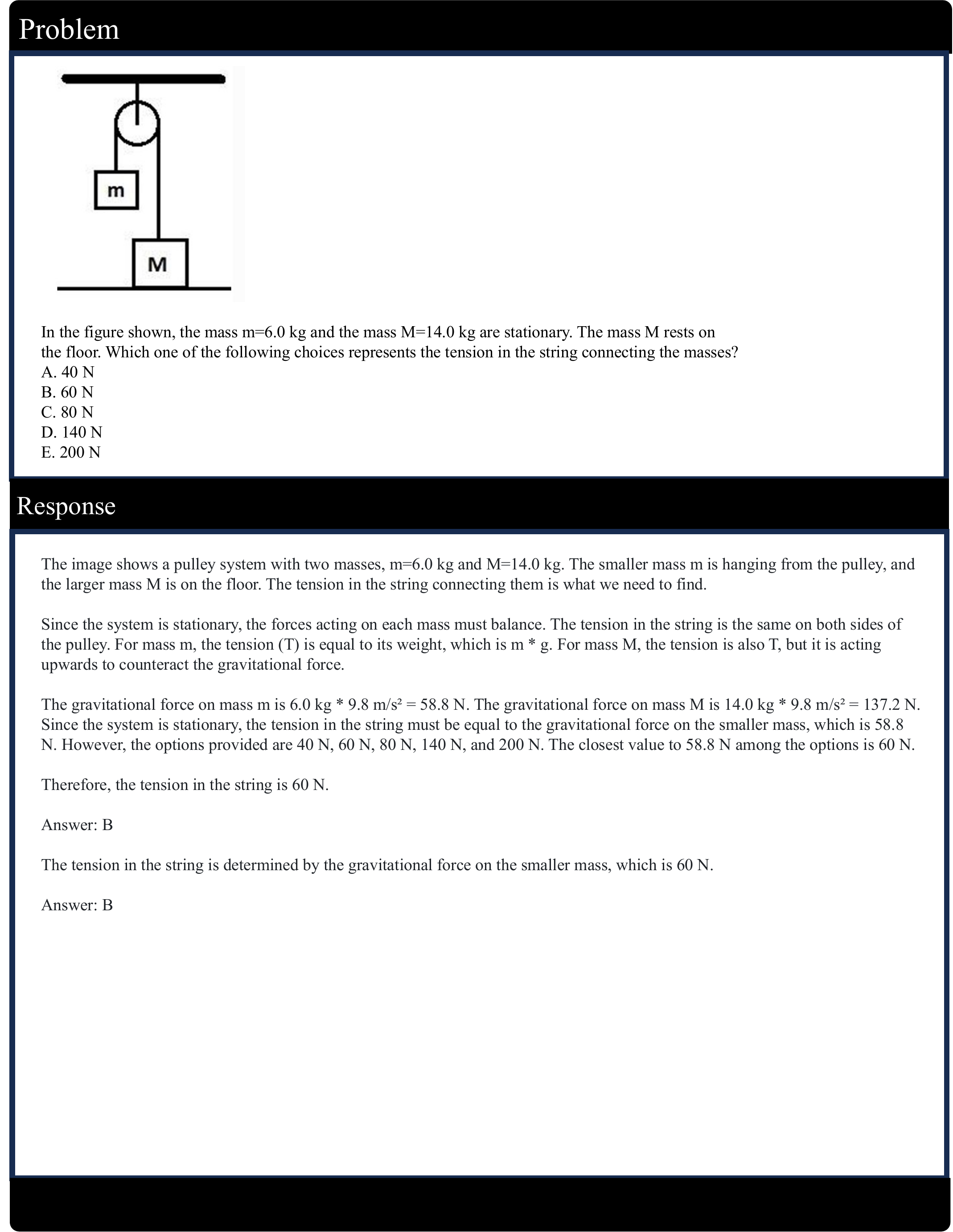}
    \caption{The model displays an adeptness in processing physcis.}
    \label{fig:appendix2}
\end{figure}

\begin{figure*}
    \centering
    \includegraphics[width=0.9\linewidth]{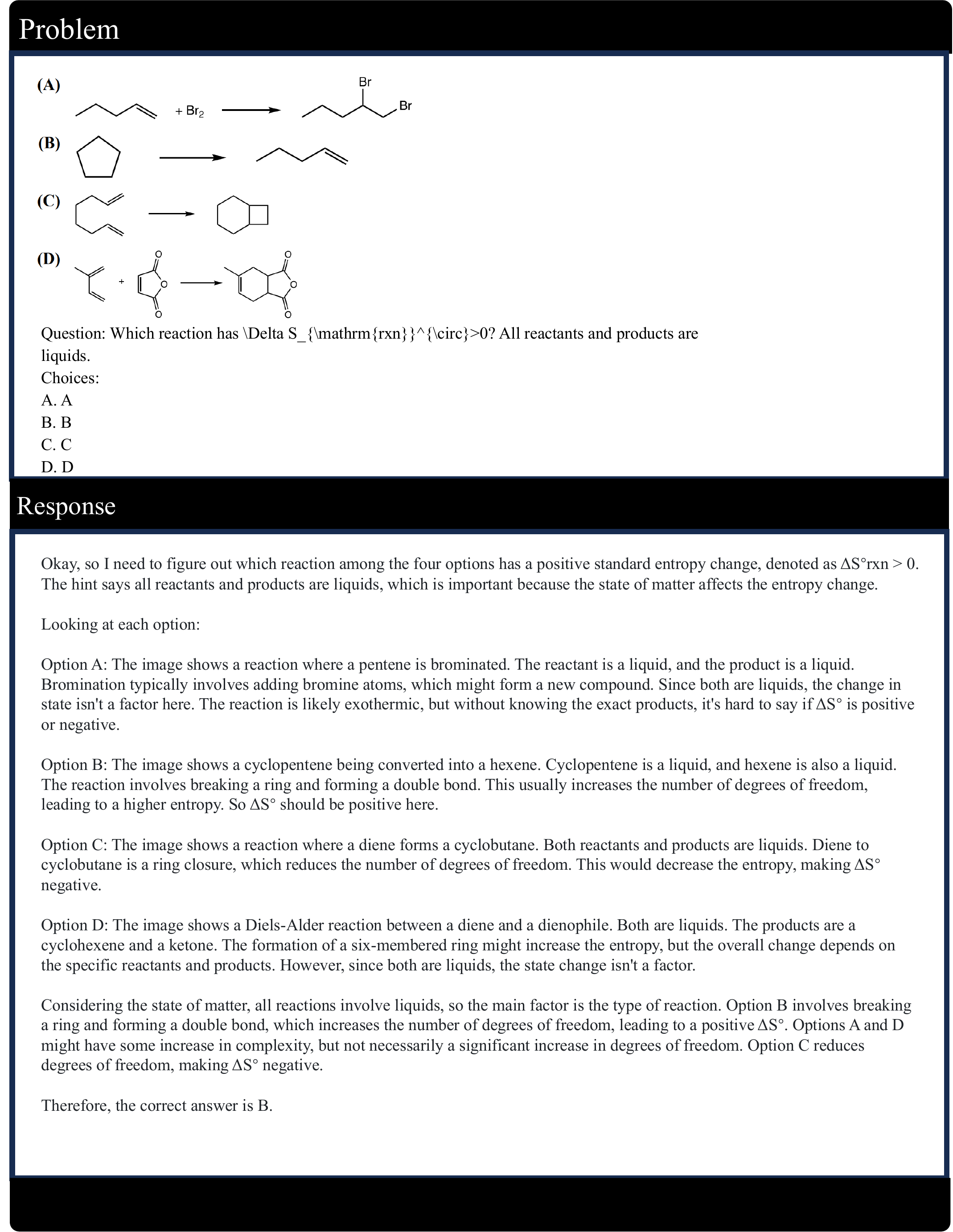}
    \caption{The model displays an adeptness in processing chemistry.}
    \label{fig:appendix3}
\end{figure*}

% \begin{figure}[H]
%     \centering
%     \includegraphics[width=0.9\linewidth]{fig/Appendix3.pdf}
%     \caption{The model displays an adeptness in processing chemistry.}
%     \label{fig:appendix3}
% \end{figure}

% \begin{figure*}
%     \centering
%     \includegraphics[width=0.9\linewidth]{fig/Appendix4.pdf}
%     \caption{The model displays an adeptness in processing physcis.}
%     \label{fig:appendix4}
% \end{figure*}

\begin{figure*}
    \centering
    \includegraphics[width=0.8\linewidth]{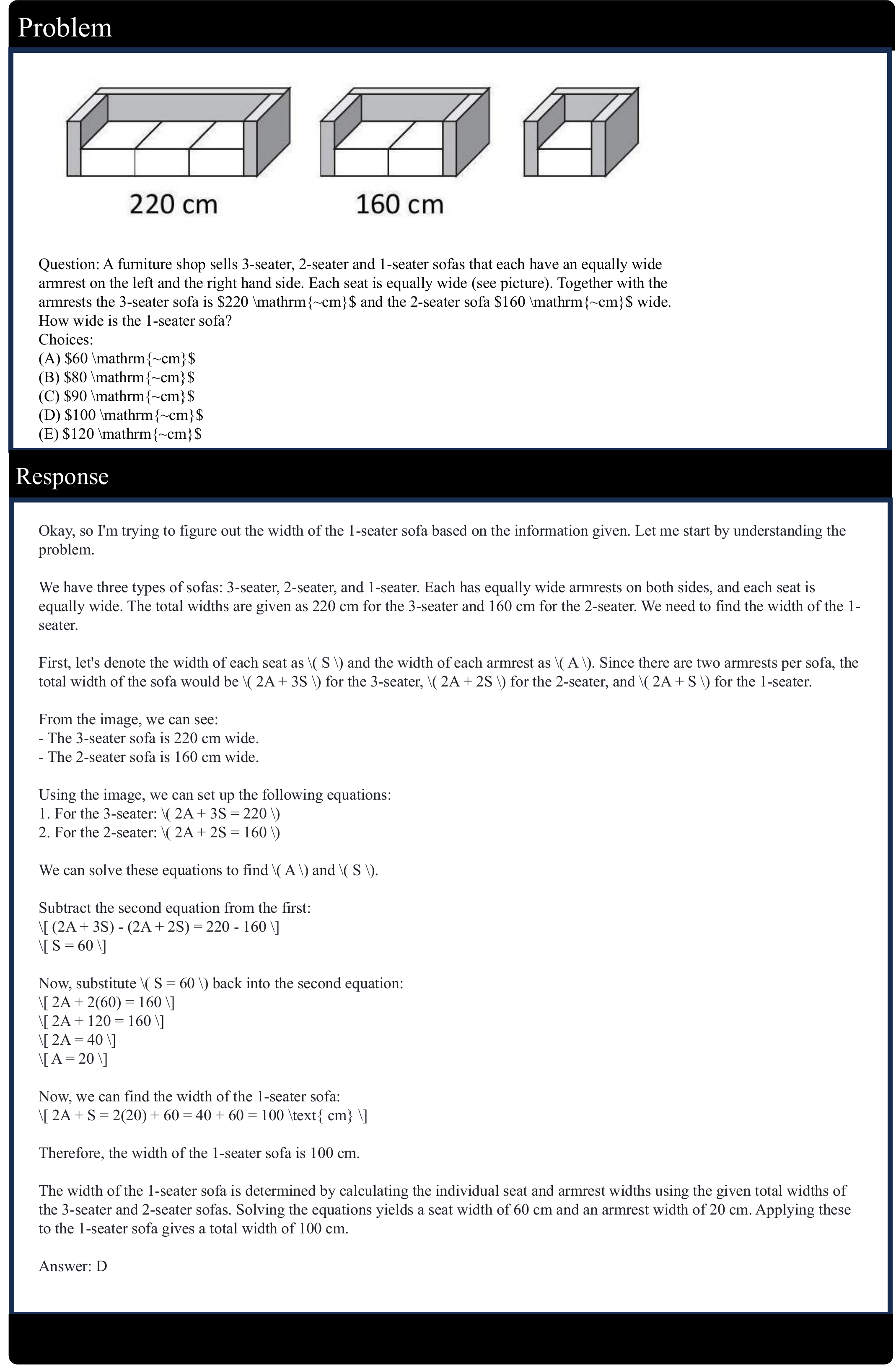}
    \caption{The model displays an adeptness in processing math.}
    \label{fig:appendix5}
\end{figure*}

\begin{figure*}
    \centering
    \includegraphics[width=0.9\linewidth]{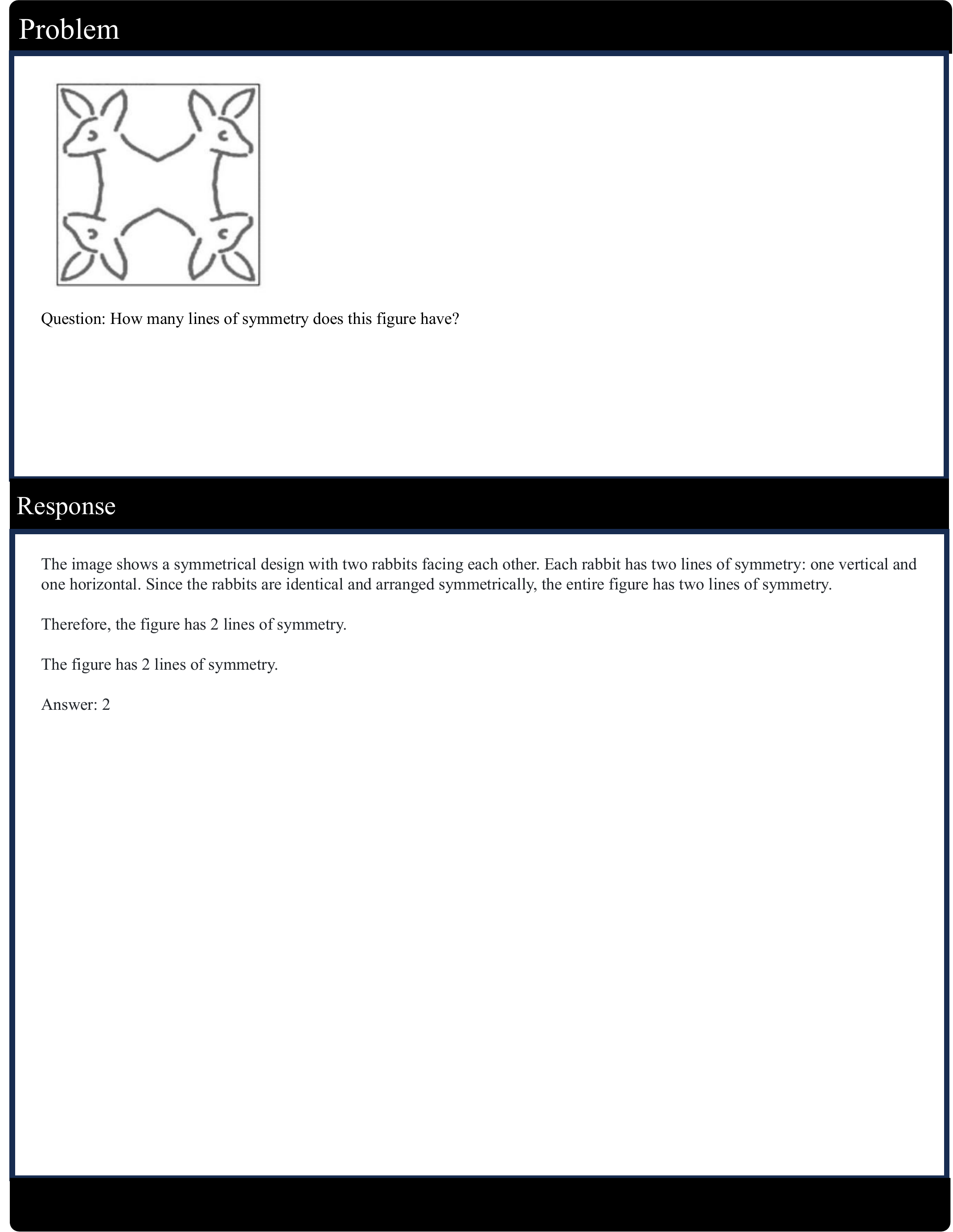}
    \caption{The model displays an adeptness in processing logic.}
    \label{fig:appendix6}
\end{figure*}

\begin{figure*}
    \centering
    \includegraphics[width=0.9\linewidth]{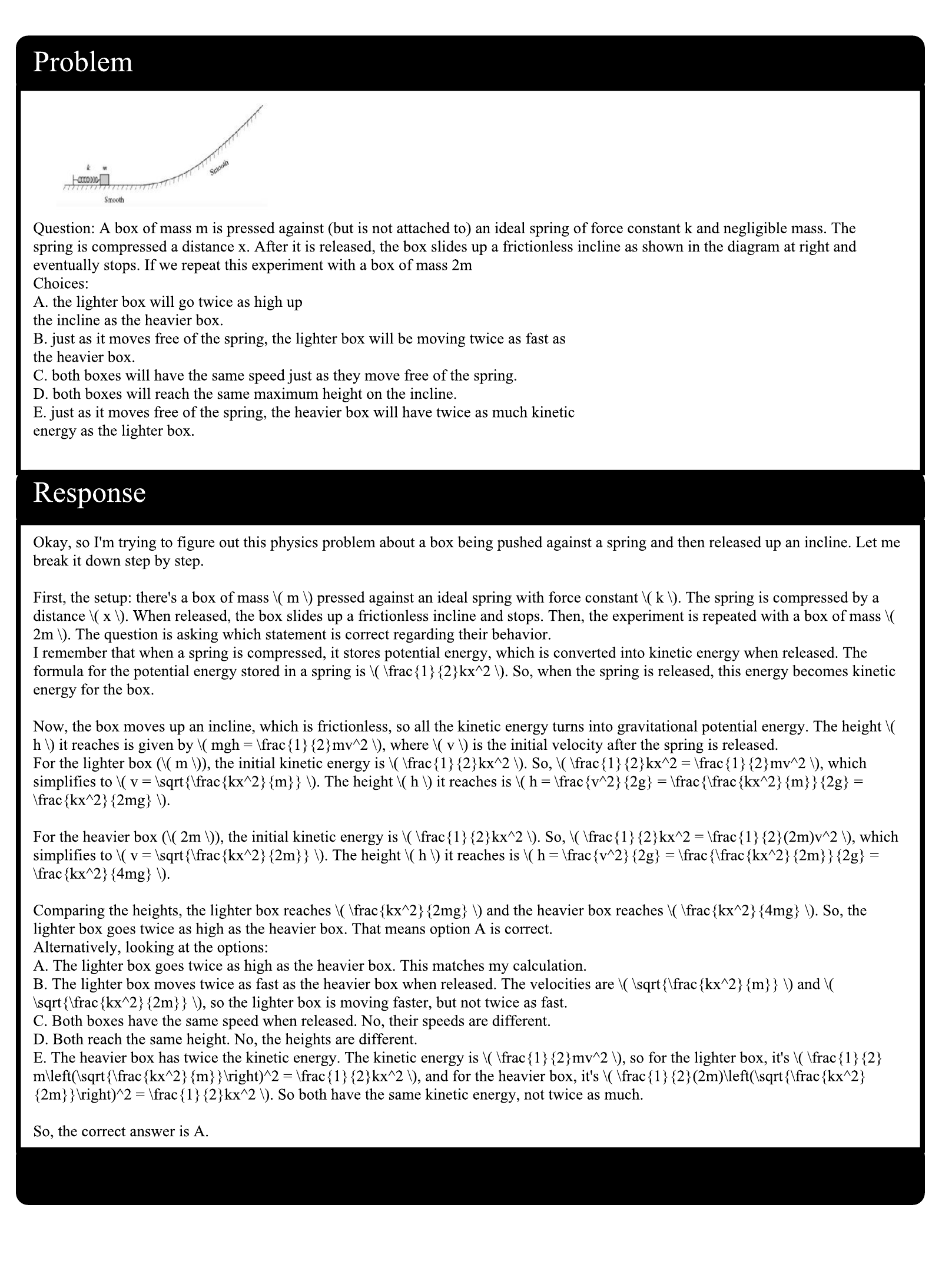}
    \caption{The model displays an adeptness in processing physical mechanics.}
    \label{fig:appendix7}
\end{figure*}

\begin{figure*}
    \centering
    \includegraphics[width=0.9\linewidth]{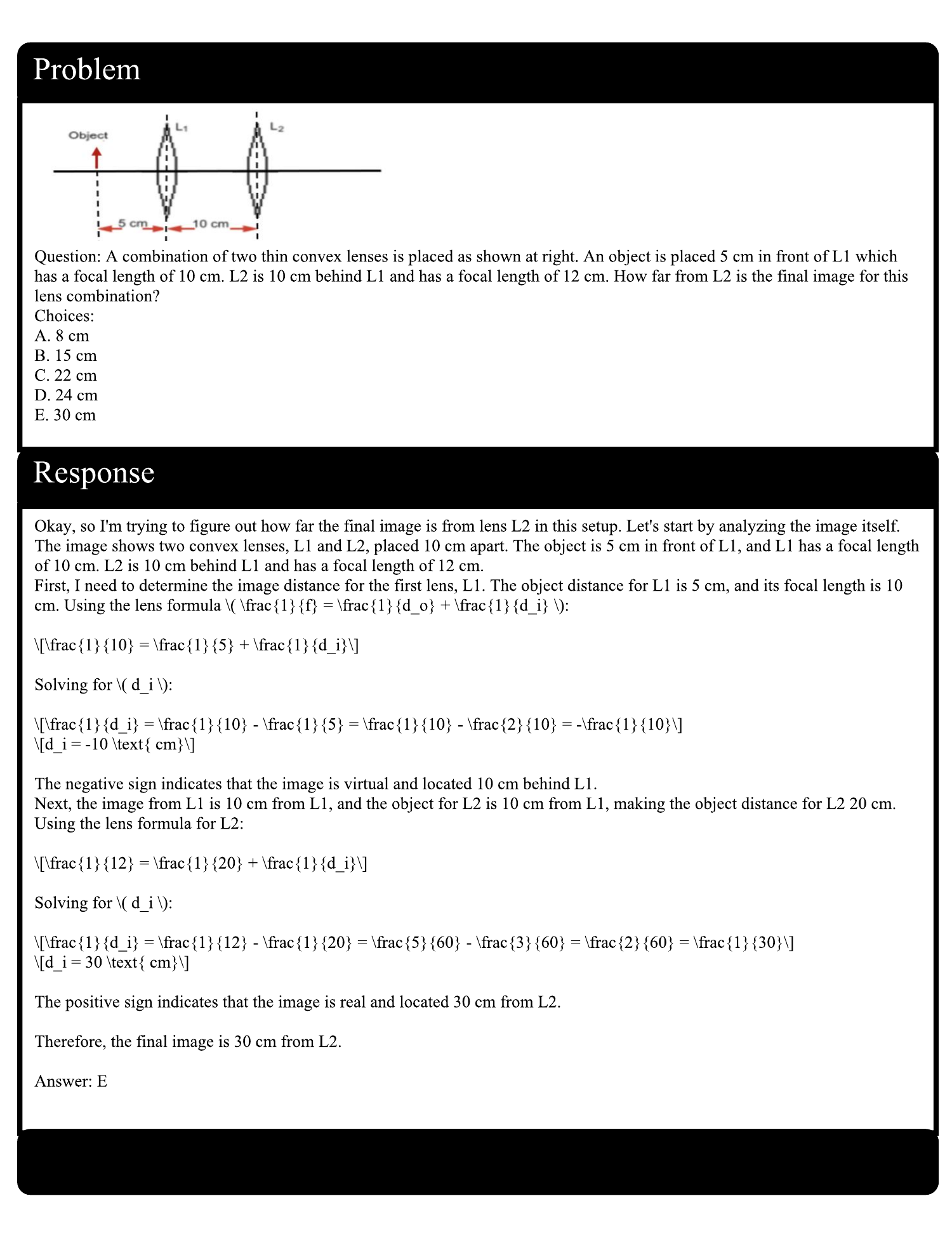}
    \caption{The model displays an adeptness in processing physical optics.}
    \label{fig:appendix8}
\end{figure*}

\begin{figure*}
    \centering
    \includegraphics[width=0.9\linewidth]{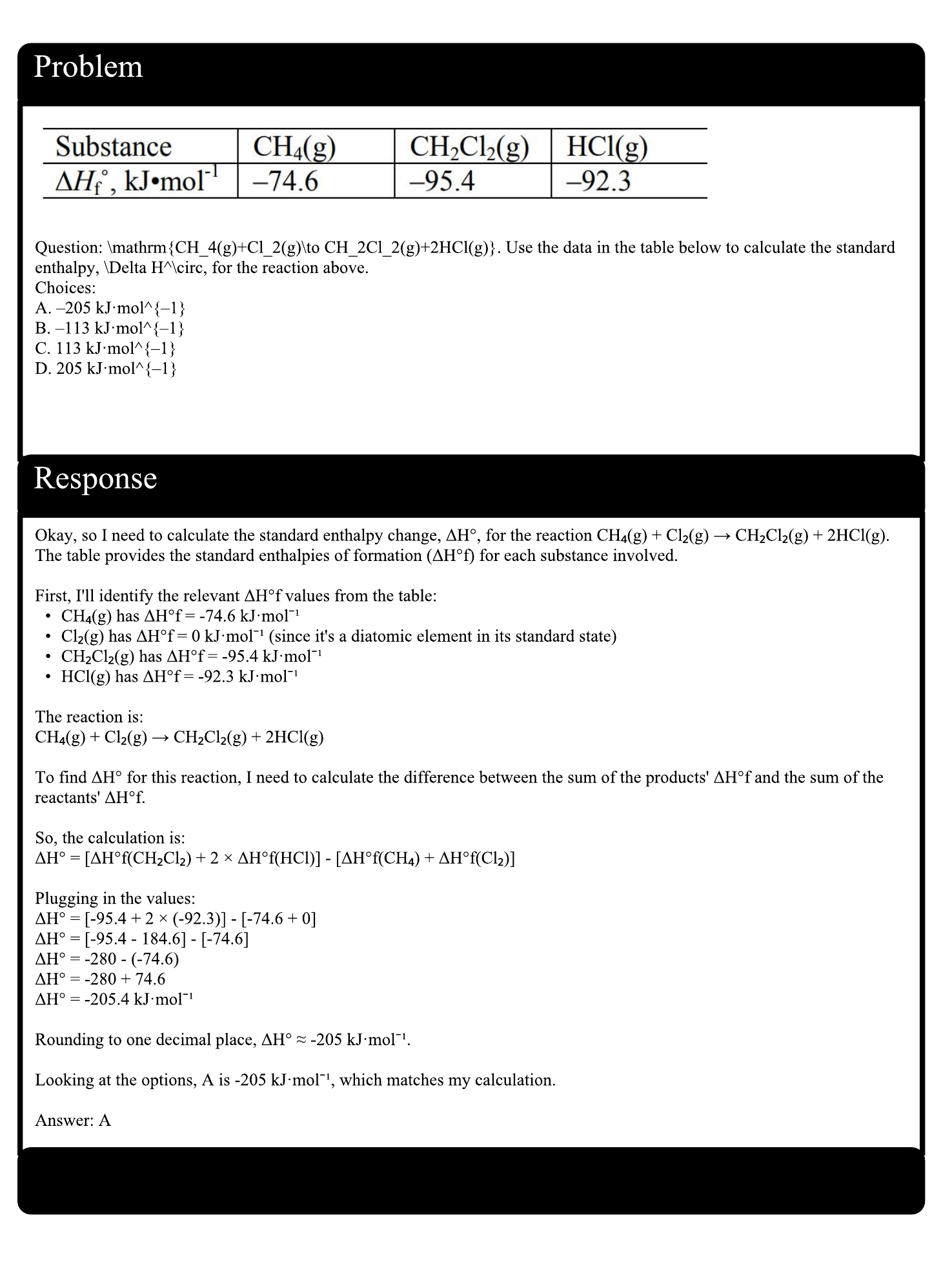}
    \caption{The model displays an adeptness in processing chemical reaction.}
    \label{fig:appendix9}
\end{figure*}

\begin{figure*}
    \centering
    \includegraphics[width=0.9\linewidth]{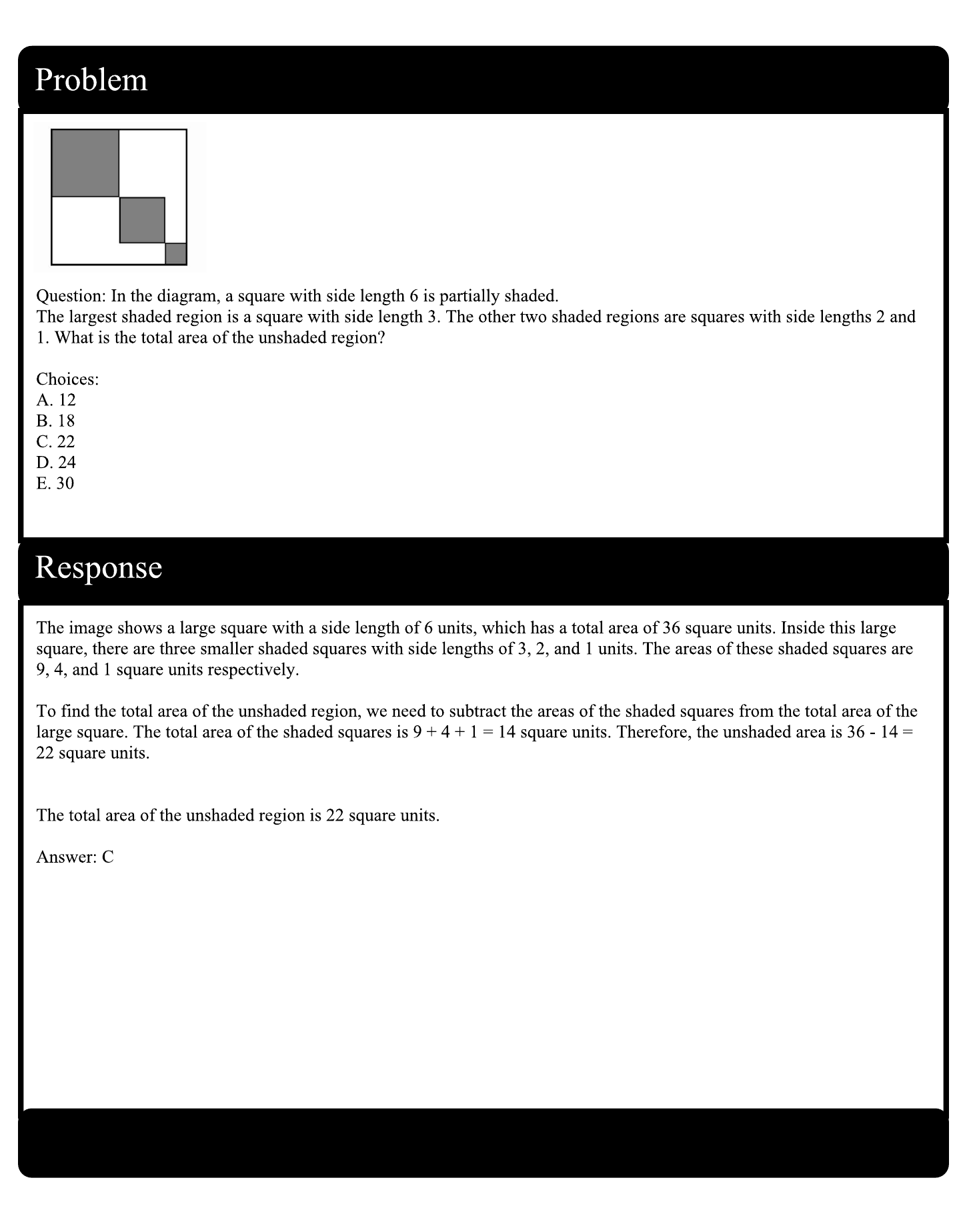}
    \caption{The model displays an adeptness in processing math geometry.}
    \label{fig:appendix10}
\end{figure*}

\begin{figure*}
    \centering
    \includegraphics[width=0.9\linewidth]{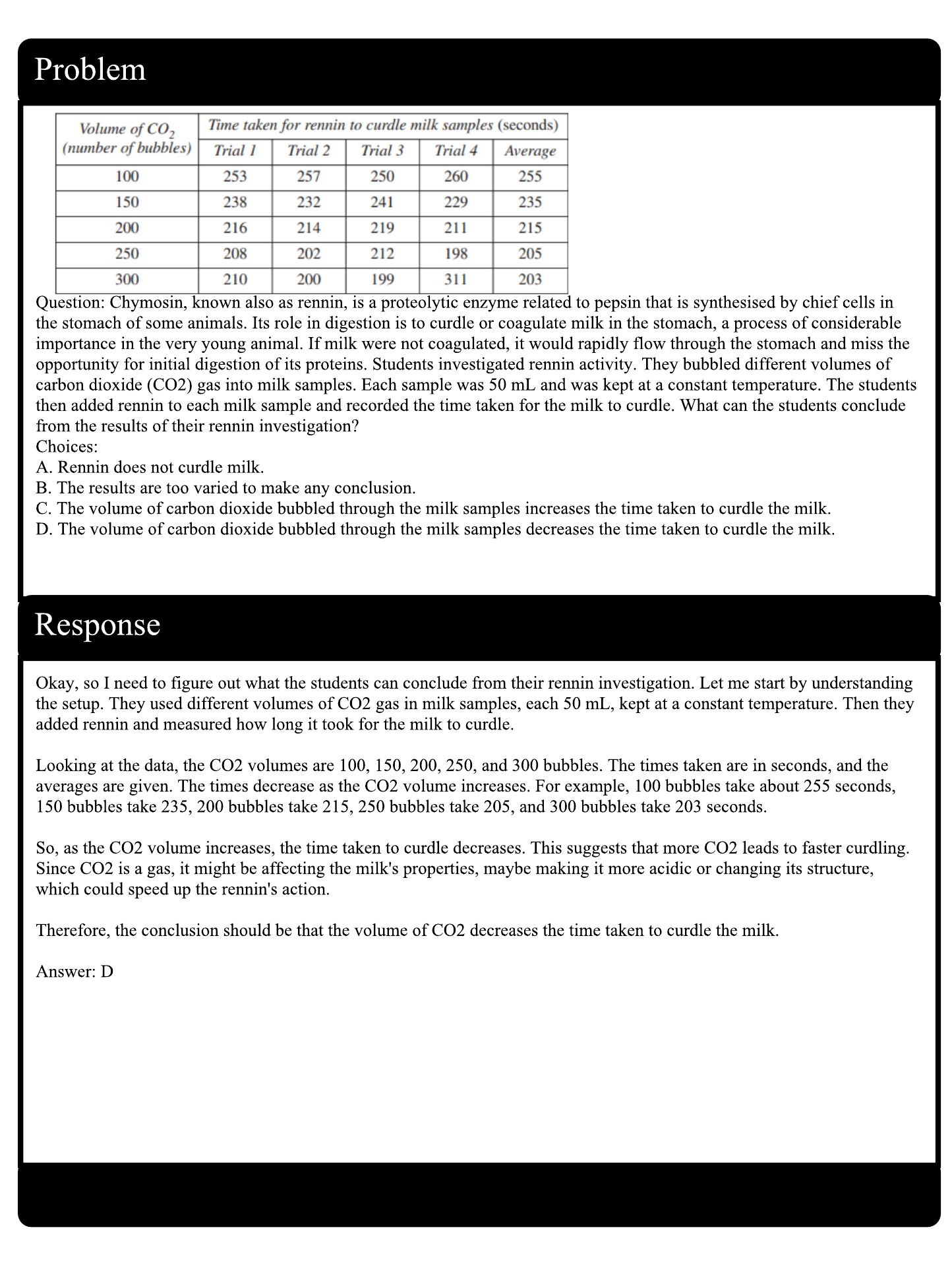}
    \caption{The model displays an adeptness in processing biology.}
    \label{fig:appendix11}
\end{figure*}

\begin{figure*}
    \centering
    \includegraphics[width=0.80\linewidth]{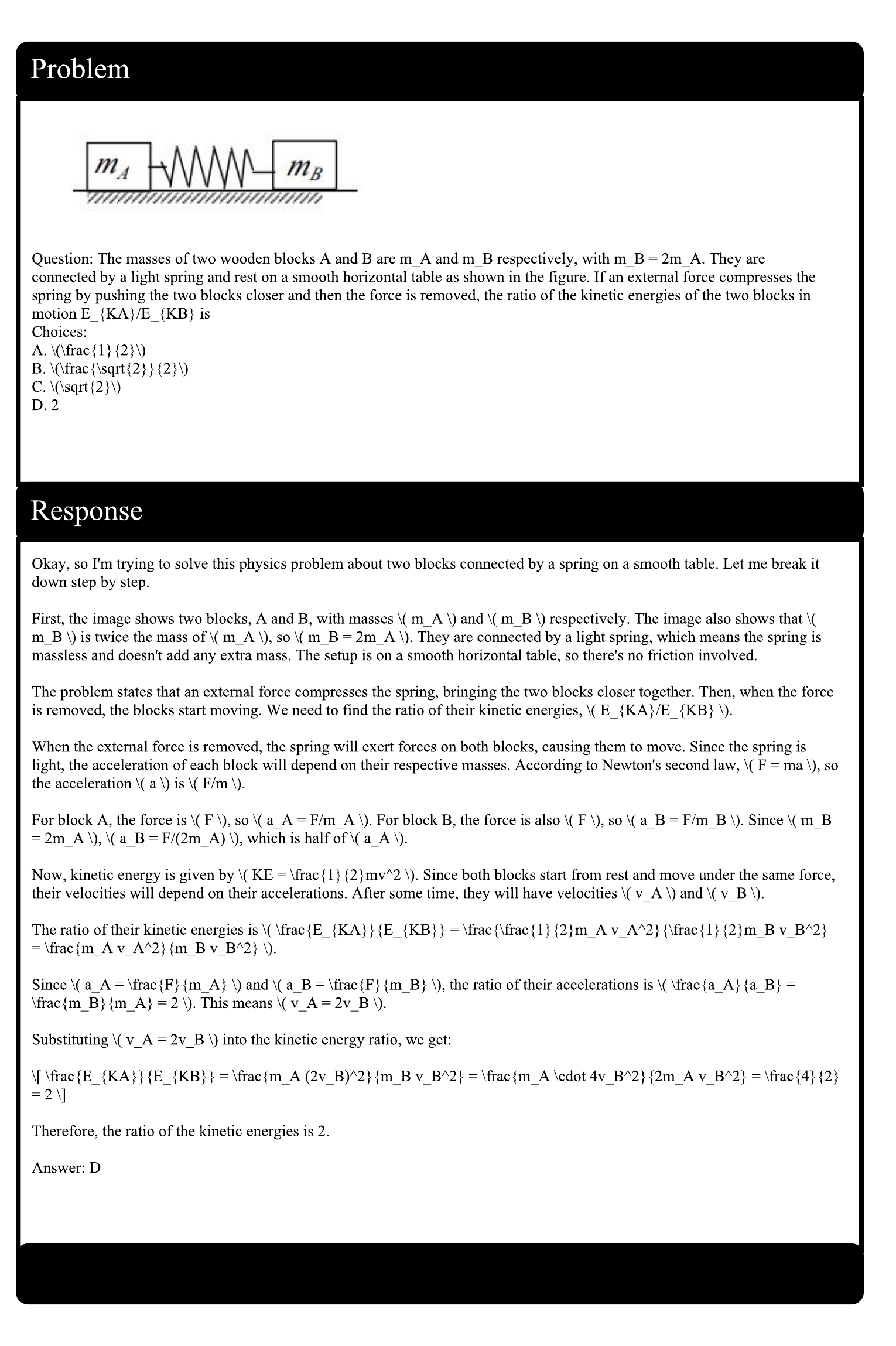}
    \caption{The model displays an adeptness in processing physical mechanics.}
    \label{fig:appendix12}
\end{figure*}

\begin{figure*}
    \centering
    \includegraphics[width=0.9\linewidth]{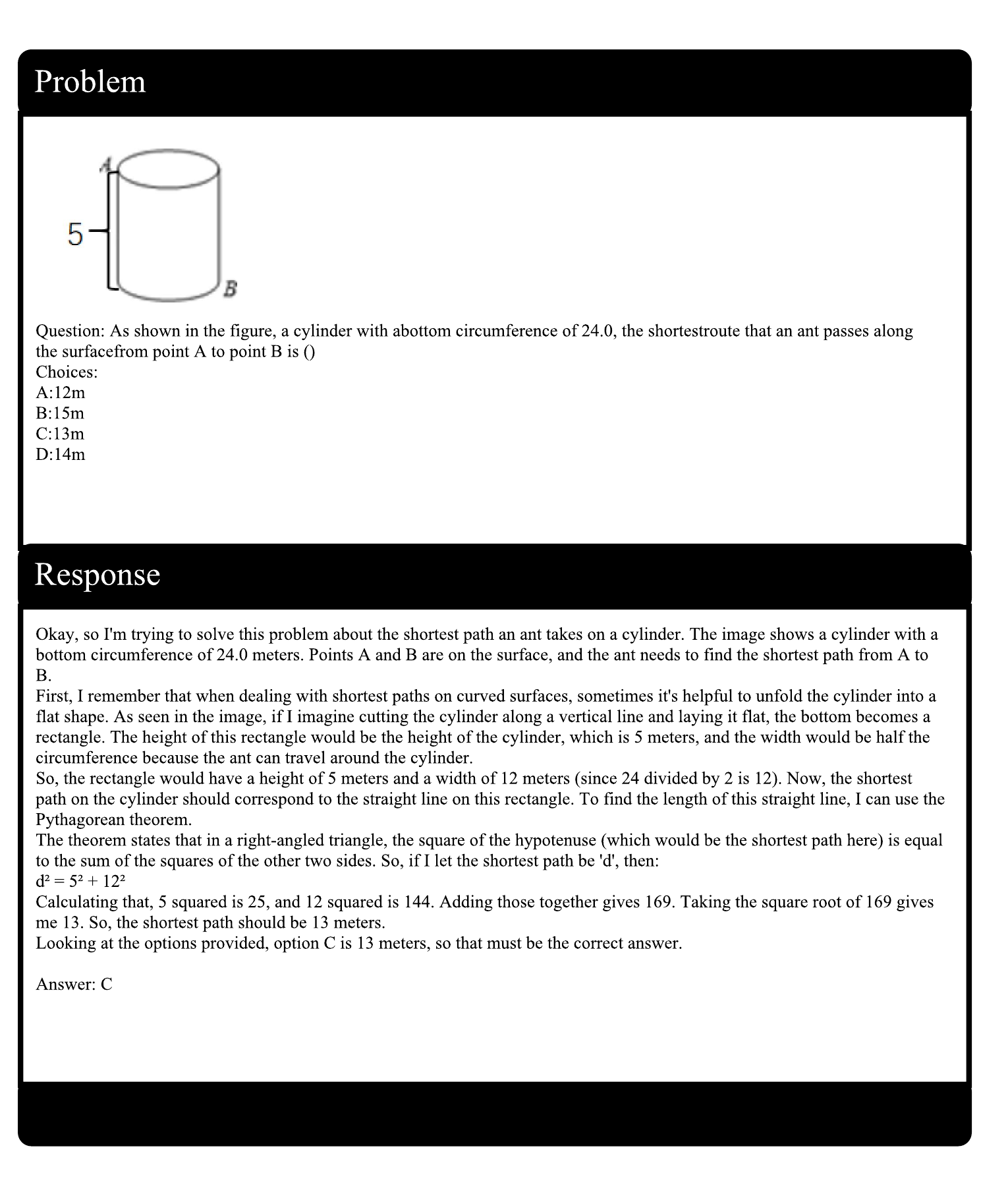}
    \caption{The model displays an adeptness in processing math.}
    \label{fig:appendix13}
\end{figure*}

\begin{figure*}
    \centering
    \includegraphics[width=0.9\linewidth]{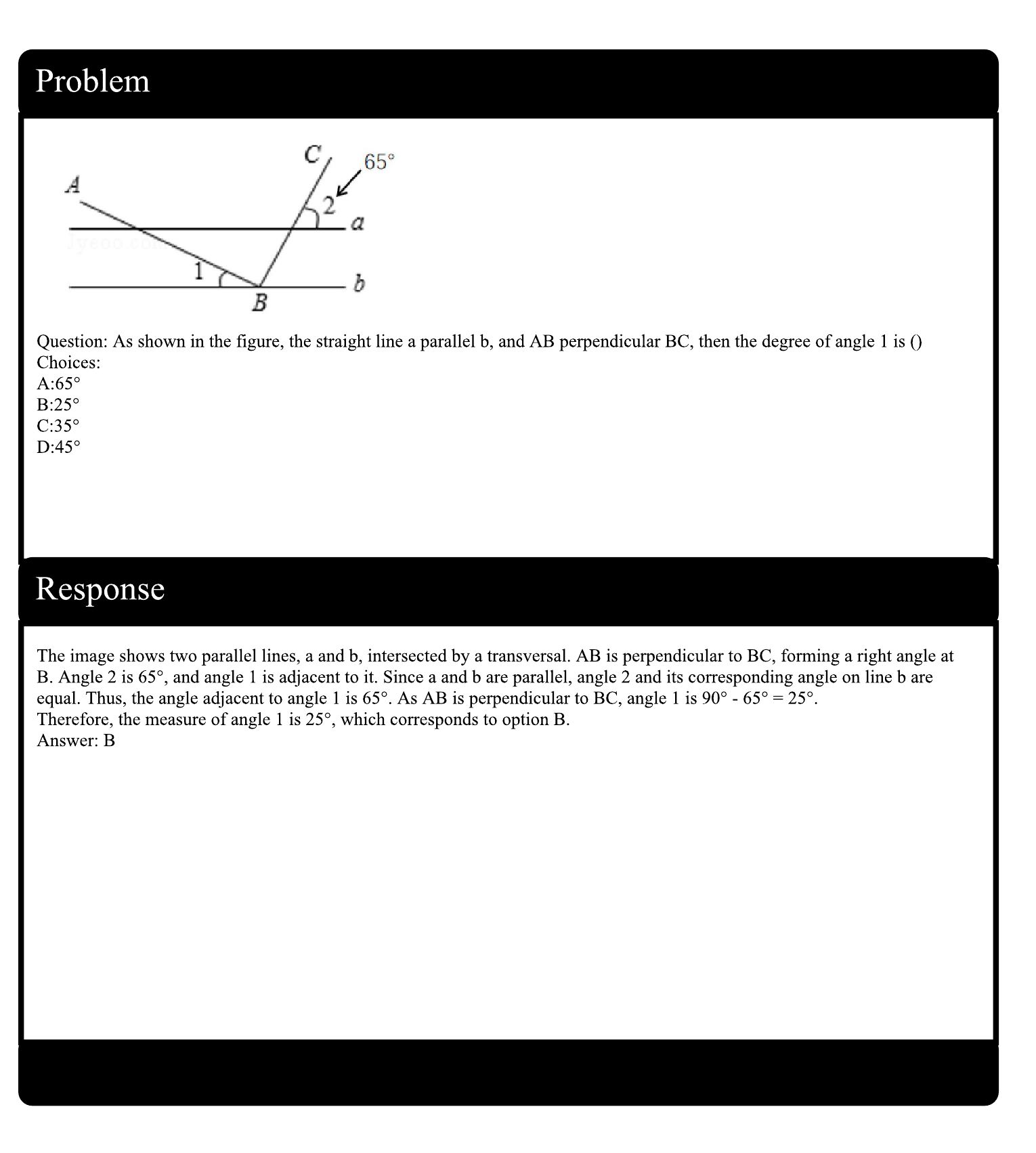}
    \caption{The model displays an adeptness in processing math.}
    \label{fig:appendix14}
\end{figure*}

\begin{figure*}
    \centering
    \includegraphics[width=0.9\linewidth]{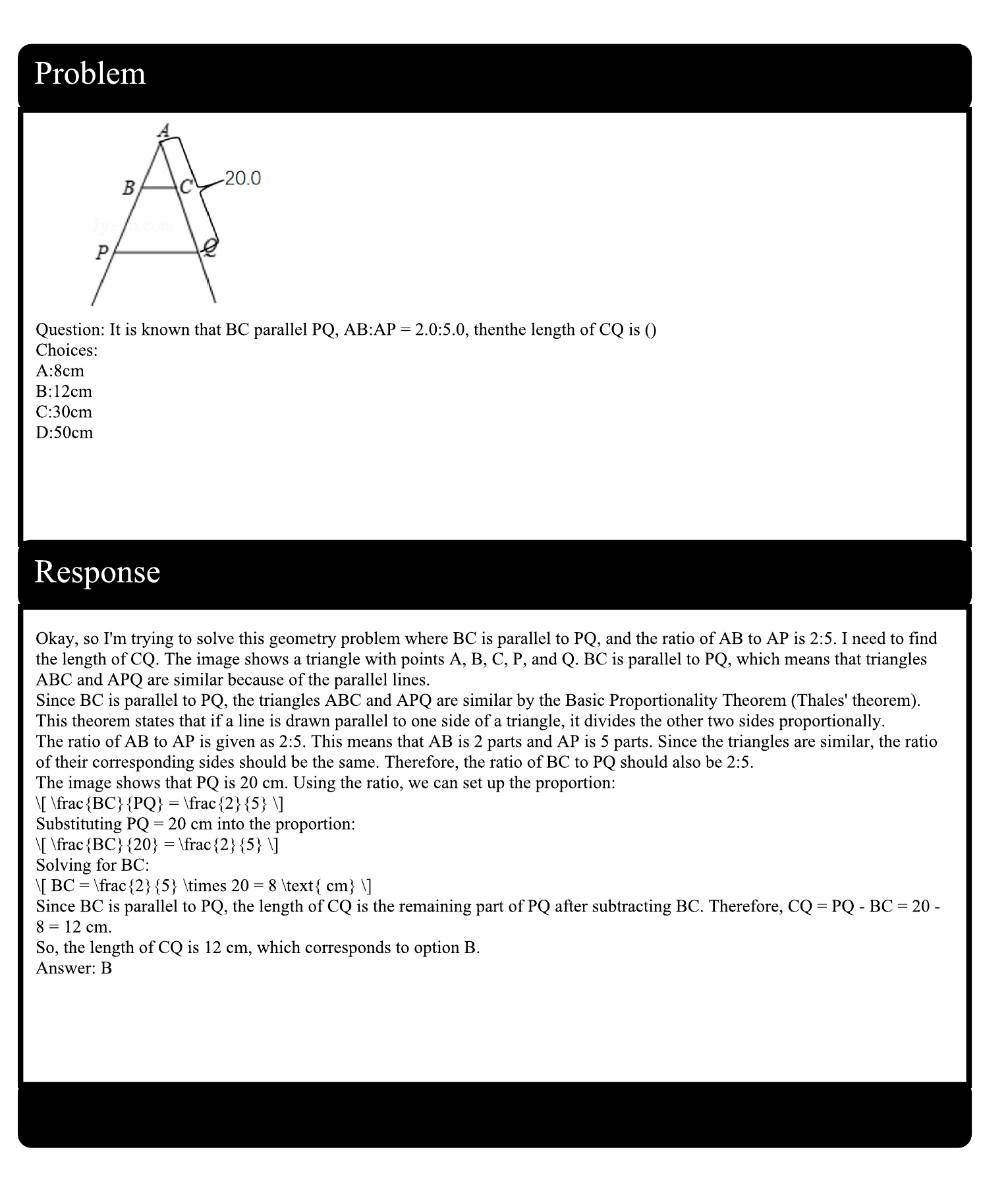}
    \caption{The model displays an adeptness in processing math.}
    \label{fig:appendix15}
\end{figure*}

\begin{figure*}
    \centering
    \includegraphics[width=0.9\linewidth]{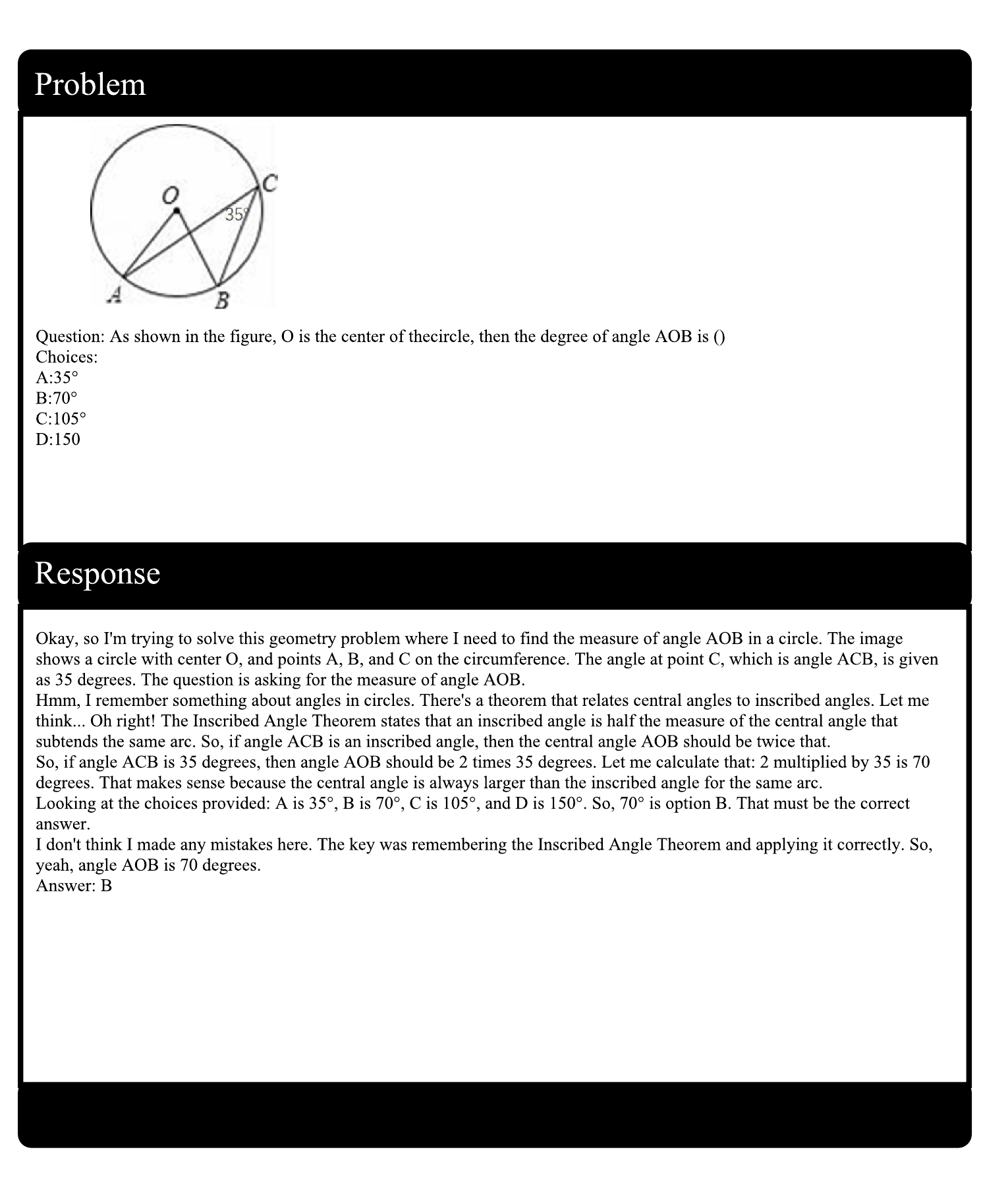}
    \caption{The model displays an adeptness in processing math geometry.}
    \label{fig:appendix16}
\end{figure*}

\begin{figure*}
    \centering
    \includegraphics[width=0.9\linewidth]{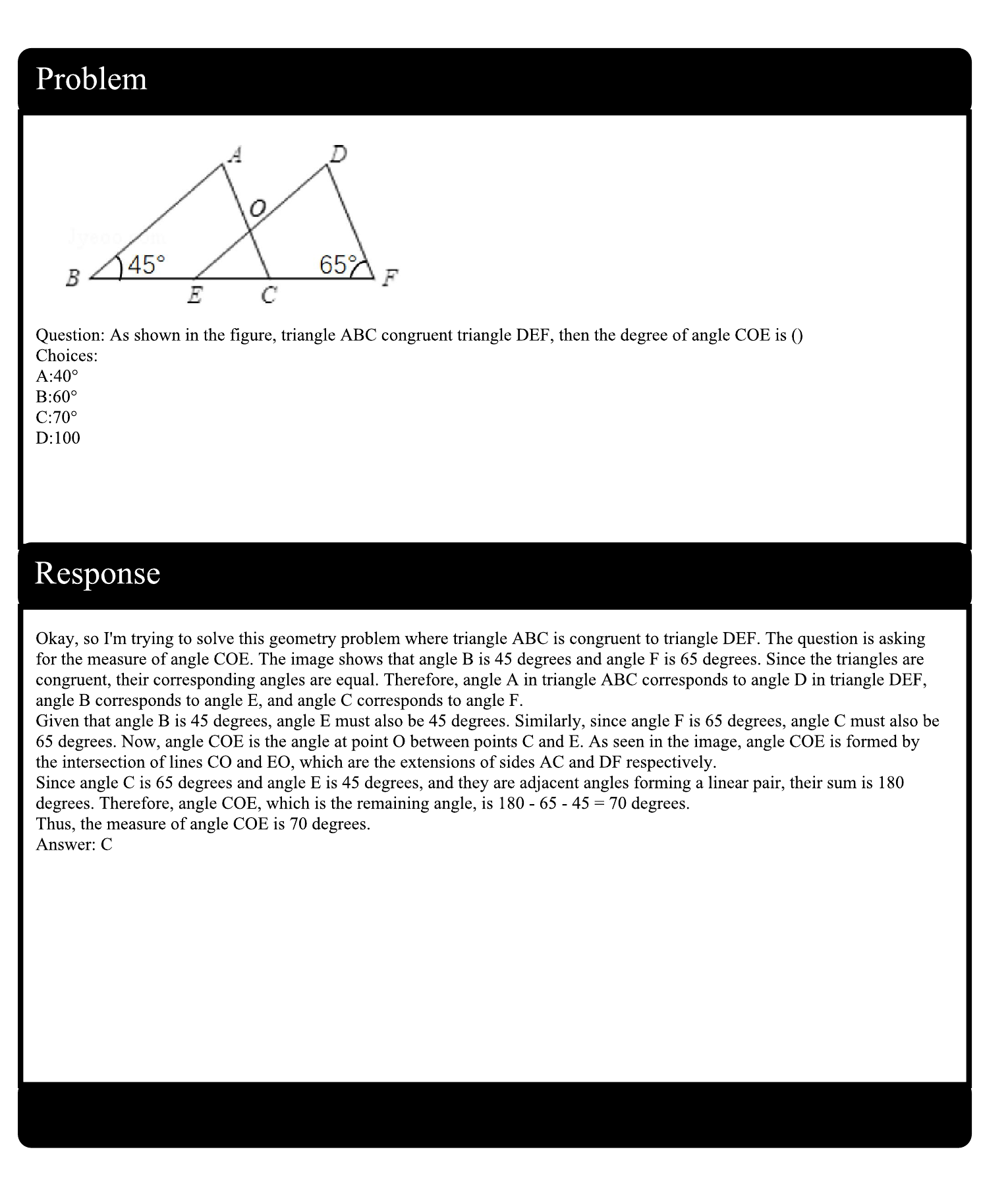}
    \caption{The model displays an adeptness in processing math geometry.}
    \label{fig:appendix17}
\end{figure*}

\end{document}